\documentclass[lettersize,journal]{IEEEtran}

\usepackage{amsmath,amsfonts}
\usepackage{algorithmic}
\usepackage{algorithm}
\usepackage{array}
\usepackage[caption=false,font=normalsize,labelfont=sf,textfont=sf]{subfig}
\usepackage{textcomp}
\usepackage{stfloats}
\usepackage{url}
\usepackage{verbatim}
\usepackage{graphicx}
\usepackage{cite}
\usepackage{algorithm}
\usepackage{algorithmic}
\usepackage{diagbox}
\usepackage{makecell}
\usepackage{enumerate}
\usepackage{setspace}
\usepackage{xcolor}
\usepackage{amsthm}
\usepackage{booktabs}
\newtheoremstyle{mystyle}
{3pt}
{3pt}
{}
{}
{\itshape}
{.}
{.5em}
{}
\theoremstyle{mystyle}
\newtheorem{assump}{Assumption}
\definecolor{lightgray}{rgb}{0.95,0.95,0.95}
\definecolor{lightblue}{rgb}{0.87,0.92,0.97}
\definecolor{lightyellow}{rgb}{1.0,0.976,0.6}
\definecolor{lightgreen}{rgb}{0.77,0.93,0.66}
\usepackage{soul}
\DeclareRobustCommand{\hlgray}[1]{{\sethlcolor{lightgray}\hl{#1}}}
\DeclareRobustCommand{\hlblue}[1]{{\sethlcolor{lightblue}\hl{#1}}}
\DeclareRobustCommand{\hlyellow}[1]{{\sethlcolor{lightyellow}\hl{#1}}}
\DeclareRobustCommand{\hlgreen}[1]{{\sethlcolor{lightgreen}\hl{#1}}}

\begin{document}

\title{An Embeddable Implicit IUVD Representation for Part-based 3D Human Surface Reconstruction}

\author{
Baoxing Li,
Yong Deng,
Yehui Yang,
Xu Zhao*,~\IEEEmembership{Member,~IEEE}
\thanks{* Xu Zhao is the corresponding author.}
\\
{Department of Automation, Shanghai Jiao Tong University}
}


\maketitle

\begin{abstract}
To reconstruct a 3D human surface from a single image, it is crucial to simultaneously consider human pose, shape, and clothing details. Recent approaches have combined parametric body models (such as SMPL), which capture body pose and shape priors, with neural implicit functions that flexibly learn clothing details. However, this combined representation introduces additional computation, e.g. signed distance calculation in 3D body feature extraction, leading to redundancy in the implicit query-and-infer process and failing to preserve the underlying body shape prior. To address these issues, we propose a novel \textit{IUVD-Feedback} representation, consisting of an \textit{IUVD occupancy function} and a \textit{feedback query algorithm}. This representation replaces the time-consuming signed distance calculation with a simple linear transformation in the \textit{IUVD space}, leveraging the SMPL UV maps. Additionally, it reduces redundant query points through a feedback mechanism, leading to more reasonable 3D body features and more effective query points, thereby preserving the parametric body prior. Moreover, the IUVD-Feedback representation can be embedded into any existing implicit human reconstruction pipeline without requiring modifications to the trained neural networks. Experiments on the THuman2.0 dataset demonstrate that the proposed IUVD-Feedback representation improves the robustness of results and achieves three times faster acceleration in the query-and-infer process. Furthermore, this representation holds potential for generative applications by leveraging its inherent semantic information from the parametric body model.
\end{abstract}

\begin{IEEEkeywords}
3D Human Surface Reconstruction, Implicit Representation, UV Map, Human Body Prior, Acceleration
\end{IEEEkeywords}

\section{Introduction} \label{sec:introduction}

Reconstructing 3D human surface from color images is useful in many applications as mixed reality, film making, virtual try-on and so forth. To date it is still an open problem due to the myriad varieties in human pose, shape and clothing details. To model these varieties, an appropriate 3D human representation is critical. The \textit{parametric body models} \cite{anguelov2005scape,loper2015smpl,pavlakos2019expressive,osman2022supr}, composed by deformable triangle meshes and learnable parameters, are convenient to represent human pose and shape, but lack clothing details. The \textit{neural implicit functions} \cite{mescheder2019occupancy,park2019deepsdf,saito2019pifu,saito2020pifuhd} can reconstruct clothed humans by indicating whether a space point is inside the human surface \cite{mescheder2019occupancy} or by inferring the signed distance between a space point and the human surface \cite{park2019deepsdf}, which is called the \textit{query-and-infer} process. The implicit functions perform well in learning the geometric details of human surface but struggle to keep the body prior.

\begin{figure}[!tp]
    \centering
    \includegraphics[width=0.99\linewidth]{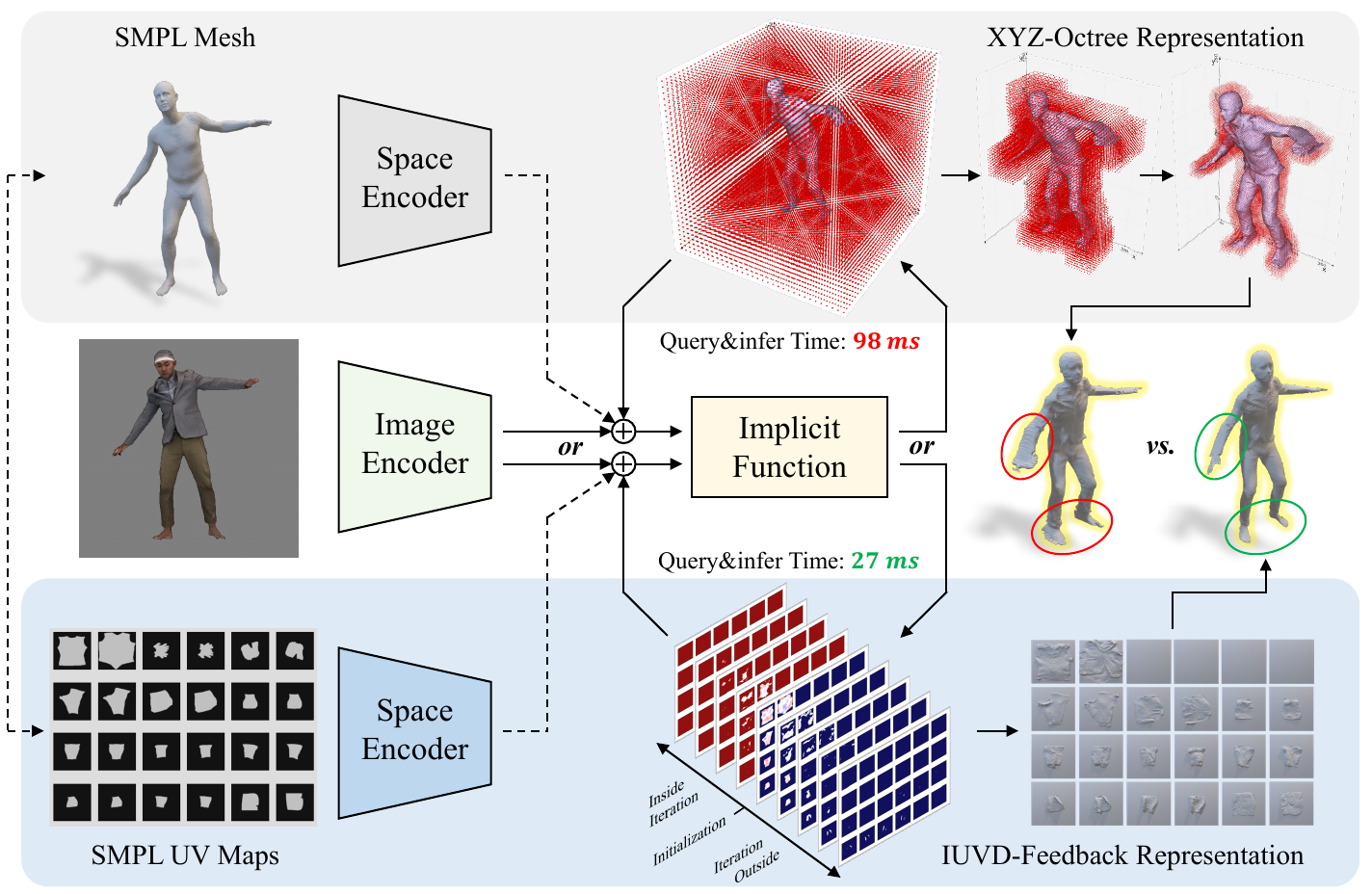}
    \caption{Comparing the pipelines of implicit human surface reconstruction in the traditional \hlgray{XYZ space} and in the proposed \hlblue{IUVD space}, there are three advantages of the proposed IUVD-Feedback representation: 1) It accelerates the query\&infer and visualization steps, by replacing the redundant SDF calculation and reducing the complexity of marching cubes respectively. 2) It preserves more robust topology of human surface, thus preventing non-human shapes. 3) It produces semantic-aware results, which enables part-based surface editing. Note that the encoders and implicit function modules are replaceable and the space encoders are optional in the pipeline.}
    \label{fig:intro_abstract}
\end{figure}

To combine the merits of parametric body models and neural implicit functions, several approaches \cite{zheng2021pamir,zheng2021deepmulticap,xiu2022icon} have been proposed. Their core ideas can be summarized as a \textit{space encoder}, as shown in Fig. \ref{fig:intro_abstract}, that takes an estimated parametric body model, e.g. SMPL \cite{loper2015smpl}, as input, and outputs a vector of body features for each space point, as an complement of the image-based features. The combined representations have improved the accuracy of reconstruction results, however, they also introduce additional computation, which may exacerbate the redundancy of the implicit query-and-infer process and reduce the completeness of the underlying body shape prior.

In fact, the problems of the combined representation are caused by two more fundamental problems. One is the compatibility between parametric body models and neural implicit functions, the other is the redundancy of neural implicit functions. 
\textbf{1) Compatibility problem.} It is not trivial to encode the dynamic parametric body mesh into an implicit function because of the gap between the explicit and implicit representations. Taking \cite{xiu2022icon} as an example, it is time-consuming to determine the \textit{signed distance field} (SDF) between the parametric body mesh and space points. This problem has been attacked by replacing the explicit parametric mesh with neural implicit representations \cite{deng2020nasa,alldieck2021imghum,mihajlovic2022coap}. But these models are unable to reconstruct clothed humans. 
\textbf{2) Redundancy problem.} The implicit query-and-infer process requires massive computation to evaluate the query points far from the human surface, which actually have little contribution to the final result. To reduce this redundancy, Li et al. \cite{li2020monocular} design an octree-based surface localization algorithm. It successfully eliminates the unnecessary query points, but can not guarantee the robustness of results due to the lack of a human shape prior.

\begin{figure*}[!tp]
    \centering
    \includegraphics[width=0.95\linewidth]{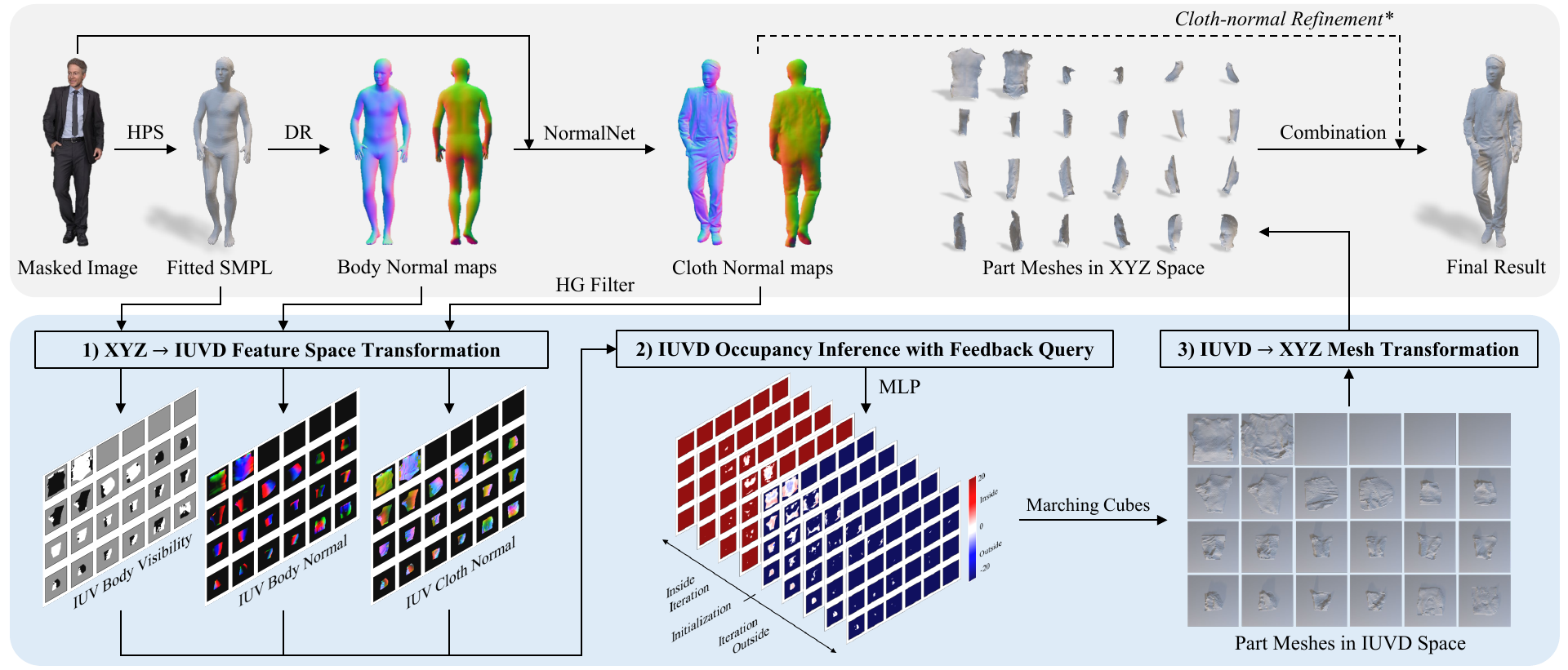}
    \caption{Overview of 3D human surface reconstruction with the proposed IUVD-Feedback representation. Given a masked color image, we first obtain the SMPL mesh, body normal maps and cloth normal maps where HPS denotes \textit{Human Pose and Shape estimation} and DR denotes \textit{Differentiable Rendering}. Then the implicit 3D human surface reconstruction process \cite{xiu2022icon} is restructured and accelerated in \hlblue{IUVD space} (Sec. \ref{sec:iuvd_representation}) by three steps: 1) XYZ to IUVD feature space transformation (Sec. \ref{sec:iuvd_feature}), 2) IUVD occupancy inference with feedback query (Sec. \ref{sec:iuvd_query}), and 3) IUVD to XYZ mesh transformation (Sec. \ref{sec:iuvd_mesh}). Finally, the part-based meshes are combined and optionally refined in XYZ space as the reconstruction result.}
    \label{fig:intro_overview}
\end{figure*}

To solve the above two problems simultaneously, we exploit the relationships between parametric body models and neural implicit functions. 
First, compared to calculating the SDF value given a set of query points and a dynamic body mesh, it is much easier to generate the query points based on the parametric body mesh. 
Second, the query points near the body mesh are usually more effective for the final visualization, which deserve more attention than those far from the mesh. 
Inspired by the above observations, we replace the traditional XYZ space with an \textit{IUVD space} based on the unwarped UV maps \cite{guler2018densepose} of the SMPL model, where $I=24$ means twenty-four indexed body parts, \textit{U} and \textit{V} denote the axes of the 2D texture for part-based SMPL meshes, and \textit{D} is a clothing deformation axis. In this space, the clothed human is represented by an implicit \textit{IUVD occupancy function}.

In the IUVD space, the compatibility and redundancy problems are solved as following. 
1) For the compatibility problem, since the SMPL mesh is unwarped into UV maps and the part-based meshes are aligned with the IUVD occupancy function, the SDF information is locally decoupled from the pose and shape information. Therefore, the time-consuming SDF calculation can be replaced with a linear transformation, where the query points are generated around the SMPL mesh and the SDF values are simply determined by the \textit{D}-coordinates. 
2) For the redundancy problem, the IUVD space can easily exclude the points far from the human surface, thus eliminating most of the unnecessary query points. To this end, we design a novel \textit{feedback query algorithm}. In this algorithm, the query points are initialized on the SMPL surface and evaluated iteratively. Taking the previous inference as a feedback, the next batch of query points is generated inside or outside of the SMPL surface along the surface normal directions. By assuming the continuity of the clothed human surface, more than half of the query points can be reduced in this process. 
By solving these two fundamental problems, the redundancy of the implicit query-and-infer process is largely minimized and the underlying human shape prior is successfully preserved.

The proposed IUVD occupancy function, as well as the feedback query algorithm, collectively called the \textit{IUVD-Feedback} representation, can be embedded into existing implicit 3D human surface reconstruction approaches, such as PIFu \cite{saito2019pifu}, PaMIR \cite{zheng2021pamir} and ICON \cite{xiu2022icon}. Fig. \ref{fig:intro_abstract} abstracts the pipelines of these approaches, where the image encoder, space encoder, and implicit function are parameterized by different neural networks. Experiments prove that by replacing the traditional XYZ-Octree representation with the IUVD-Feedback representation, the efficiency and robustness of the reconstruction can be improved. Besides, since the semantic information of the parametric body model is fully succeeded by the reconstructed 3D model, it has potential to be used in part-based human surface editing applications.

Fig. \ref{fig:intro_overview} shows the usage of the IUVD-Feedback representation in ICON \cite{xiu2022icon}. Given a masked color image of a clothed human, the fitted SMPL model is first estimated by \cite{zhang2021pymaf} and then rendered to obtain the front and back body normal maps. The input image and body normal maps are fed into a NormalNet \cite{saito2020pifuhd} to predict the front and back cloth normal maps. Based on these features, the IUVD-Feedback representation is embedded into the implicit reconstruction pipeline by the following three steps. Firstly, the extracted features are transformed from the original XYZ space to IUVD space by UV mapping \cite{blinn1976texture}, where a set of \textit{convex assumptions} is introduced to ensure the equivalence of the feature transformation. Secondly, the proposed feedback query algorithm is used to accelerate the query-and-infer process in IUVD space, where the implicit function does not need to be re-trained. Finally, the body part meshes are separately extracted in IUVD space using marching cubes \cite{newman2006survey} and then transformed back into XYZ space, which is faster than previous surface localization algorithms. Besides, a cloth-normal refinement step used in \cite{xiu2022icon} is optionally used to obtain more details on the reconstructed surface. To sum up, the contribution of this paper is three fold.
\begin{enumerate}
    \item A new implicit 3D human representation, IUVD occupancy function, is presented in this paper. This is a general-purpose representation with significant potential to be embedded into existing implicit human surface reconstruction pipelines.
    \item A novel feedback query algorithm for clothed human surface localization is designed in IUVD space, which reduces more redundancy in implicit human reconstruction than existing octree-based algorithm.
    \item Experiments show that the proposed IUVD-Feedback representation accelerates the query-and-infer process by three more times than \cite{xiu2022icon}, and improves the robustness of results without re-training the neural networks.
\end{enumerate}

\section{Related Work} \label{sec:related_work}

3D human surface reconstruction has been an active research topic for over two decades. We review the approaches related to parametric body models and neural implicit functions, which are the cornerstones of this research.

\begin{table*}[!tp]
    \centering
    \caption{Comparison between the 2D UV mapping based 3D learnable representations.}
    \resizebox{0.99\linewidth}{!}{
        \begin{tabular}{llll}
        \toprule
        Representation & Creation of UV maps & General pipeline for 3D surface reconstruction or rendering & Ex~/~Im \\
        \midrule
        SMPL-based IUV map \cite{wang2023learning,zhang2023pymafx} & DensePose UV maps \cite{guler2018densepose} & RGB image $\rightarrow$ IUV map in image sapce $\rightarrow$ SMPL mesh & Explicit \\
        Extrapolated IUV map \cite{xie2022temporaluv,jafarian2023normalguided} & DensePose UV maps + extrapolation \cite{xie2022temporaluv} & RGB image $\rightarrow$ IUV map (+ geometric property) $\rightarrow$ Re-textured image & Explicit \\
        UV volumes \cite{chen2023uv} & DensePose UV maps \cite{guler2018densepose} & RGB image $\rightarrow$ UV volumes in XYZ space $\rightarrow$ IUV map $\rightarrow$ Novel view image & Implicit \\
        SMPL-based UV+D \cite{alldieck2018detailed,lazova2019360degree,alldieck2019tex2shape} & Single joint SMPL UV parameterization \cite{alldieck2019tex2shape} & RGB image $\rightarrow$ UV displacement map $\rightarrow$ SMPL+D mesh & Explicit \\
        Geometry image \cite{sinha2016deep,sinha2017surfnet,zhang2022reggeonet,zhang2023flatteningnet} & Authalic UV parametrization \cite{sinha2016deep} & RGB image / Point cloud $\rightarrow$ UV map + geometric property $\rightarrow$ XYZ mesh & Explicit \\
        IUVD occupancy function (Ours) & DensePose UV maps \cite{guler2018densepose} & RGB image $\rightarrow$ IUVD features $\rightarrow$ IUVD occupancy $\rightarrow$ UVD meshes $\rightarrow$ XYZ mesh & Implicit \\
        \bottomrule
        \end{tabular}
    }
    \label{tab:uv_representations}
\end{table*}

\subsection{Parametric Human Body Recovery}

To represent the 3D human body, statistical body models \cite{anguelov2005scape,loper2015smpl,pavlakos2019expressive,osman2022supr} are learned from 3D scans and motion capture data \cite{mahmood2019amass}, thus carrying a robust prior of human pose and shape. These models are parameterized to represent the dynamic human body with an animatable triangle mesh. The pose and shape parameters can be estimated from color images using optimization-based \cite{bogo2016keep,lassner2017unite,fan2021revitalizing} or regression-based \cite{kanazawa2018endtoend,kolotouros2019learning,kocabas2021pare} methods. Nowadays, the parametric body models have been applied to more complex problems such as temporal human tracking \cite{kocabas2020vibe}, occluded human estimation \cite{khirodkar2022occluded}, multi-person reconstruction \cite{choi2022learning}, and expressive body recovery \cite{feng2021collaborative}. Although the parametric body models can only represent naked bodies, they provide a strong prior of human pose and shape for clothed human surface reconstruction, which is a prerequisite of our method.

\subsection{SMPL-based Human Surface Reconstruction}

The SMPL \cite{loper2015smpl} model is one of the most popular parametric body models. However, it is unable to represent clothing details. Alldieck et al. \cite{alldieck2018video} propose to add offsets to the SMPL template mesh for surface deformation, which is called the SMPL+D model. This model has been widely used in clothed human reconstruction \cite{moon20223d,zhao2022highfidelity}, which enriches the clothing details of the SMPL model. But the results are restricted to a fixed topology \cite{zhu2019detailed} or limited by clothing types \cite{bhatnagar2019multigarment}. Xiu et al. \cite{xiu2023econ} overcome these defects by integrating the SMPL-X \cite{pavlakos2019expressive} model into the pipeline of normal integration, thus reconstructing more realistic details of loose clothing from the predicted normal maps. But the iterative optimization process is time-consuming.

\subsection{UV-based Human Surface Reconstruction}

The SMPL surface can be unwarped onto 2D image plane by UV mapping \cite{blinn1976texture}. The unwarped UV maps contain the surface topology information and redefine the task of 3D human surface reconstruction into two paradigms.

\textit{1) UV coordinates estimation.} By estimating the SMPL UV coordinates and body part indices of image pixels, i.e. the IUV map defined by DensePose \cite{guler2018densepose}, the SMPL mesh can be indirectly reconstructed from the input image \cite{wang2023learning,zhang2023pymafx}. To capture more details, the part-based UV maps can be extrapolated to fit the silhouette of loose clothing, such as dress \cite{xie2022temporaluv}. However, these extrapolated UV maps are limited to specific clothing types and are primarily used for image re-texturing \cite{jafarian2023normalguided}. Similarly, Chen et al. \cite{chen2023uv} propose UV volumes to implicitly generate IUV map for novel views, enabling free-viewpoint rendering applications, though not used for geometric reconstruction.

\textit{2) UV displacement estimation.} Estimating the offsets of the vertices of SMPL mesh is equivalent to estimating a displacement UV map. Based on this idea, Alldieck et al. \cite{alldieck2018detailed} transform visible texture and the predicted segmentations from image space into UV space to estimate the displacement UV map. Lazova et al. \cite{lazova2019360degree} propose to first complete the partially estimated segmentations and visible textures in UV space, then estimate the displacement UV map to create a fullly-textured 3D avatar. Tex2Shape \cite{alldieck2019tex2shape} firstly unwarps the input image into UV space, then estimates the normal map and displacement map in UV space, and finally generates an SMPL+D mesh. More generally, without using the SMPL model, some methods directly estimate geometry images, i.e. the 3D coordinates of UV maps, using geometric properties, such as curvature, from RGB images \cite{sinha2016deep,sinha2017surfnet} or from point clouds \cite{zhang2022reggeonet,zhang2023flatteningnet}. These UV maps can be unwarped from the 3D mesh of any simple object via authalic surface parametrization but they lack the pose and shape prior information specific to 3D humans provided by SMPL.

The differences between these UV-based representations are summarized in Table \ref{tab:uv_representations}. Unlike existing methods, the proposed IUVD representation integrates the SMPL UV maps into the implicit human surface reconstruction pipeline, offering a flexible, detailed and time-efficient solution.

\subsection{Implicit Human Surface Reconstruction}

In addition to the surface-based methods, the 3D human surface can also be reconstructed using volume-based approaches that are not limited by fixed topology. However, explicit voxel-based methods \cite{varol2018bodynet,zheng2019deephuman} are limited by the large memory cost. In recent years, implicit reconstruction methods have been proposed to solve this problem. Based on learnable implicit functions, such as occupancy functions \cite{saito2019pifu,saito2020pifuhd,chan2022integratedpifu} and signed distance functions \cite{onizuka2020tetratsdf,jiang2022selfrecon,alldieck2022photorealistic}, usually parameterized by Multi-Layer Perceptrons (MLPs), the 3D human surface can be reconstructed with unlimited resolution through a memory-efficient query-and-infer process. But their results are sometimes unrealistic or even incomplete due to the lack of human pose and shape priors. Consequently, the parametric body models, e.g. SMPL, have been utilized to extract additional 3D body features \cite{zheng2021pamir,xiu2022icon,chan2022spifu,yang2023dif,liao2023highfidelity,zhang2024sifu} as a complement to the 2D image features. For example, Zheng et al. \cite{zheng2021pamir} use a 3D encoder to convert the SMPL model into a 3D feature volume. But the global encoding manner is hard to be generalized to out-of-distribution human poses. Therefore, Xiu et al. \cite{xiu2022icon} replace the global body feature with signed distance values as a local body feature, whose results are more robust to pose variations. In this way, the parametric body models successfully preserve the pose and shape priors in implicit human surface reconstruction.

\subsection{Speeding-up Implicit Surface Reconstruction}

Although the neural implicit functions are memory-efficient, the query-and-infer process is time-consuming especially when the space resolution is high. Existing researches have proved that redundancy exists in this process. To solve the redundancy, Li et al. \cite{li2020monocular} design an octree-based coarse-to-fine strategy to reduce the query points far from the reconstructed surface. However, the results may be inaccurate if the segmentation step fails. Feng et al. \cite{feng2022fof} propose to represent a 3D human with a Fourier occupancy field. By discarding the tail terms of the Fourier series, it successfully reduces the redundancy of useless high frequency components. But when the SMPL model is used as an additional input, its running speed is reduced. In contrast, we focus on reducing the redundancy in implicit 3D human surface reconstruction by fully exploiting the parametric body models.

\section{Implicit IUVD Representation} \label{sec:iuvd_representation}

To represent a 3D clothed human is to simultaneously represent the pose, shape, and clothing details of the target person. Since the parametric body models, e.g. SMPL \cite{loper2015smpl}, have learned the pose and shape information of a naked body, the remaining task is to add displacement to the body mesh, e.g. the SMPL+D \cite{alldieck2018video} model. Differently, we consider modeling the clothing deformation in an implicit manner.

\begin{figure}[!tp]
    \centering
    \includegraphics[width=0.90\linewidth]{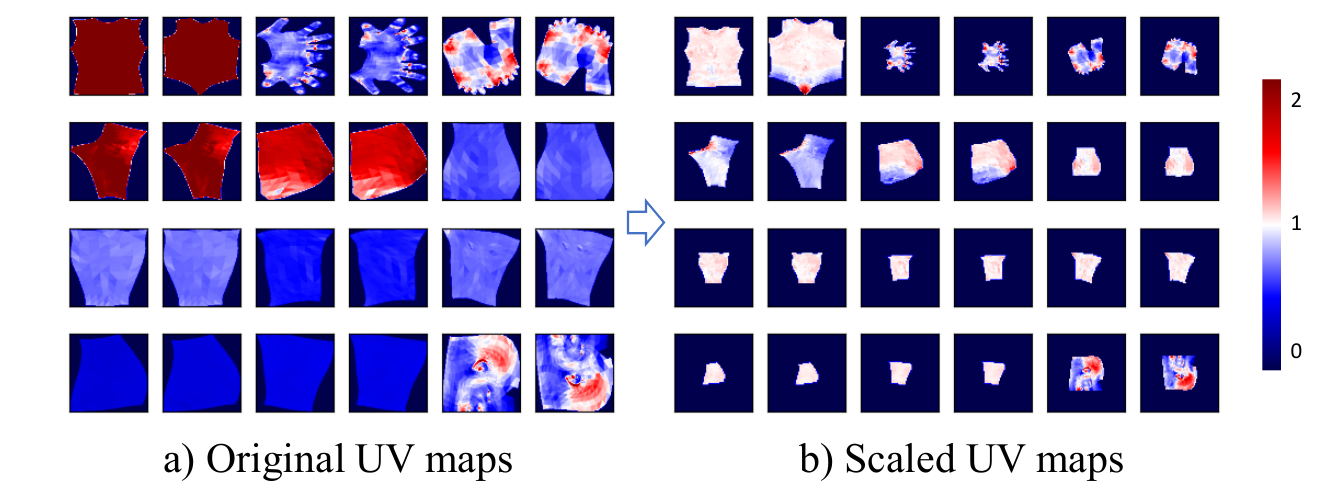}
    \vspace{-2.5mm}
    \caption{Scaling of the SMPL UV maps for nearly uniform sampling. The color indicates the proportions of different body parts. The black area is obsolete.}
    \label{fig:method_smpl_iuv_area_ratio}
    \vspace{-2.0mm}
\end{figure}

We note that the surface of SMPL model is a 2D compact smooth manifold defined in a 3D Euclidean space, ${\Phi} \subseteq \mathbb{R}^{3}$, named as the \textit{XYZ space}. And the SMPL template mesh can be unwarped onto 24 UV maps corresponding to 24 body parts defined by \cite{guler2018densepose}, as shown in Fig. \ref{fig:method_smpl_iuv_area_ratio} (a), where each pixel, with a non-zero value, of the UV maps corresponds to a 3D point of the SMPL surface. To represent the clothing details, a D-axis is then added orthogonal to the UV-axes, constructing 24 Euclidean UVD spaces, denoted as ${\Psi}_{i} \subseteq \mathbb{R}^{3}, i=1,\cdots,24$. The collection of all UVD spaces is named as the \textit{IUVD space}, $\Psi = \{\Psi_i | i=1,\cdots,I\}$, where $I=24$ is the number of body parts indexed. Note that the resolution of each UVD space is theoretically unlimited, thus satisfying the sampling scalability of a general 3D data representation \cite{feng2022fof}. The shape of the UV plane is formulated by the SMPL UV maps. And the range of the D-axis is limited to $(D_{min},D_{max})$, which will be discussed in Sec. \ref{sec:iuvd_feature}.

To ensure an even density of sampling points in each UVD space, the SMPL UV maps, denoted as $\tilde{M}$, are scaled to preserve the real proportions of different body parts. Let $\bar{A}_{xyz,i}$ denote the average area of the triangle mesh of the $i$-th body part in XYZ space, and $\bar{A}_{uv,i}$ denote the average area of the projected triangles on the $i$-th UV map. The ratio of the average areas is given by $r_i = \bar{A}_{xyz,i} / \bar{A}_{uv,i}$. Then the UV coordinates of the projected triangles are updated using Eq. \eqref{eq:uv_scale}. The scaled UV maps are shown in Fig. \ref{fig:method_smpl_iuv_area_ratio} (b).
\begin{equation} \label{eq:uv_scale}
    (u,v)_i := (u,v)_i \cdot \frac{r_i}{\mbox{max} \{r_i\}},~~ i = 1, \cdots, 24.
\end{equation}

In IUVD space, the clothed human surface is represented by an implicit \textit{IUVD occupancy function}, denoted as $f$. Let $\tilde{P}(i,u,v,d) \in \Psi$ be the query point in IUVD space and ${P}(x,y,z) \in \Phi$ be the corresponding point in XYZ space. The occupancy value $f(\tilde{P})$ shows the relationship between $P$ and the clothed human surface $S^c \subseteq \Phi$ as defined in Eq. \eqref{eq:iuvd_occupancy}.
\begin{equation} \label{eq:iuvd_occupancy}
f: \Psi \rightarrow \{0,1\}.
\end{equation}
If $f(\tilde{P})=0$, the corresponding $P$ is outside the clothed human surface. And if $f(\tilde{P})=1$, $P$ is inside the surface. According to \cite{mescheder2019occupancy}, this function is approximated with a neural network, $f_{\theta}$, where $\theta$ denotes the parameters of the network, for binary classification. The decision boundary of $f_{\theta}$ implicitly represents the clothed human surface $\tilde{S^c} \subseteq \Psi$.

Meanwhile, the IUVD occupancy function naturally carries the information of parametric body model, as the $d$ coordinate of $\tilde{P}$ indicates the relationship between $P$ and the SMPL mesh of the $i$-th body part. If $d>0$, $P$ is outside the SMPL mesh; if $d<0$, $P$ is inside the SMPL mesh.

\section{IUVD-based Human Surface Reconstruction} \label{sec:iuvd_reconstruction}

Given a masked color image and a fitted SMPL model, the 3D human surface can be reconstructed through a learned IUVD occupancy function, $f_{\theta}$, which is usually parameterized by a Multi-Layer Perceptron (MLP), as shown in Eq. \eqref{eq:occupancy_infer}.
\begin{equation} \label{eq:occupancy_infer}
    f_{\theta}(\tilde{P}) = \text{MLP}(\mathcal{F}_{iuvd}(\tilde{P})), ~~ \forall \tilde{P} \in \Psi,
\end{equation}
where $\mathcal{F}_{iuvd}(\tilde{P})$ is the feature vector of $\tilde{P}$, extracted from the input image and the fitted SMPL model in IUVD space.

Embedding the above IUVD occupancy function into existing implicit human surface reconstruction pipelines requires three steps:
1) Converting the feature vectors in XYZ space to IUVD space, i.e. $\mathcal{F}_{xyz}(P) \rightarrow \mathcal{F}_{iuvd}(\tilde{P})$,
2) Inferring the occupancy value $f_{\theta}(\tilde{P})$ in IUVD space, and
3) Extracting triangle meshes in IUVD space and then transforming them back to XYZ space.

\begin{figure*}[!tp]
    \centering
    \includegraphics[width=0.95\linewidth]{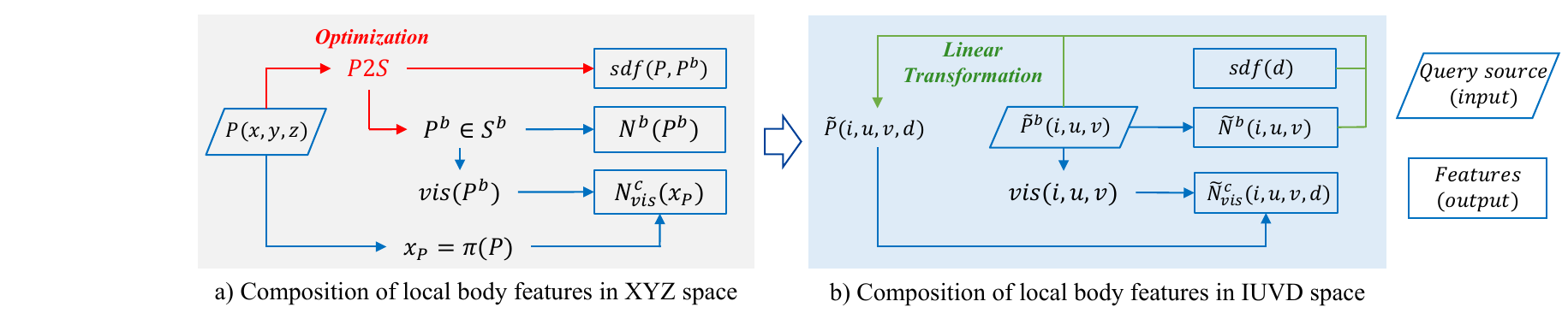}
    \caption{By changing the source of query points, the local body features in XYZ space are transformed to IUVD space. And the time-consuming P2S optimization (\textcolor{red}{red}) is replaced by a simple linear transformation (\textcolor{green}{green}) for constructing the IUVD local body features.}
    \label{fig:method_features_composition}
\end{figure*}

\subsection{XYZ to IUVD Feature Space Transformation} \label{sec:iuvd_feature}

To maintain the original performance, we do not change the feature components and the structure of any neural networks. Note that the feature vectors may differ in different implicit functions, but the approach of XYZ-IUVD feature space transformation is similar. Therefore, in this section we take ICON \cite{xiu2022icon} as an example and show the transformation of its local body features from XYZ space to IUVD space. The local body features include the SMPL body normal, the visible cloth normal, and the signed distance based on the nearest surface point of the SMPL mesh. The composition and transformation of the local body features in XYZ space and in IUVD space are discussed as follows.

\textit{1) XYZ local body features}.
As shown in Fig. \ref{fig:method_features_composition} (a), for each query point $P$ in XYZ space, the nearest point $P^b$ on the estimated SMPL surface ${S^b} \subseteq {\Phi}$, and the signed distance $sdf(P,P^b)$ is determined by solving a \textit{point-to-surface} (P2S) optimization problem \cite{zhu2010geometry}. This is a time-consuming process. Meanwhile, the projection of $P$, denoted as $x_p$, is obtained with a weak perspective camera $\pi$. The front or back cloth normal of $x_p$, predicted by a pix2pixHD \cite{wang2018highresolutiona} network (NormalNet), is selected based on the visibility of $P^b$, denoted as $N^c_{vis}(x_p)$. Finally, the XYZ local body features $\mathcal{F}_{xyz}(P) \in \mathbb{R}^7$ are defined in Eq. \eqref{eq:xyz_feature}.
\begin{equation} \label{eq:xyz_feature}
    \mathcal{F}_{xyz}(P) = [sdf(P, P^b), N^b(P^b), g(N_{vis}^c(x_p))],
\end{equation}
where $N^b(P^b) \in \mathbb{R}^3$ is the SMPL body normal of $P^b$, and $g$ is a stacked hourglass \cite{newell2016stacked} network (HGFilter).

\begin{figure}[!tp]
    \centering
    \includegraphics[width=0.90\linewidth]{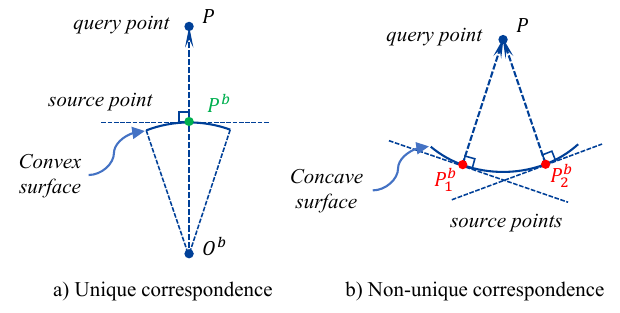}
    \caption{For a query point $P$, the corresponding source point \textcolor{green}{$P^b$} is unique when the surface is convex. But there may exist multiple source points, e.g. \textcolor{red}{$P^b_1$}, \textcolor{red}{$P^b_2$}, when the surface is concave.}
    \label{fig:method_equivalence_analysis}
\end{figure}

\textit{2) IUVD local body features}.
Firstly, by UV mapping \cite{blinn1976texture}, we build a dense correspondence between the SMPL UV maps and the attributes of SMPL mesh, including vertex position, normal orientation (\textit{IUV body normal}), and visibility information (\textit{IUV body visibility}), as visualized in Fig. \ref{fig:intro_overview}. 
Secondly, in IUVD space, we replace the P2S optimization with a simple linear transformation, as shown in Fig. \ref{fig:method_features_composition} (b), to generate the query points near the SMPL surface ${S^b}$. We define the \textit{source point} $\tilde{P}^b(i,u,v) = P^b \in S^b$ corresponding to the query points ${P} \in \Phi$ along the D-axis as shown in Fig. \ref{fig:method_equivalence_analysis} (a). Then the XYZ coordinates of $P$ are generated by the linear transformation $\mathcal{L}$ shown in Eq. \eqref{eq:pb_to_p}.
\begin{equation}\label{eq:pb_to_p}
    P(x,y,z) = \mathcal{L}(\tilde{P}) = \tilde{P}^b(i,u,v) + \tilde{N}^b(i,u,v) \cdot sdf(d),
\end{equation}
where $\tilde{N}^{s}(i,u,v) = N^b(P^b)$ is the IUV body normal, $sdf(d) = \alpha \cdot d, d \in [D_{min},D_{max}]$ is a given signed distance function, and $\alpha$ is a scale factor. 
Finally, based on this linear transformation, the IUVD local body features $\mathcal{F}_{iuvd}(\tilde{P}) \in \mathbb{R}^7$ are derived as Eq. \eqref{eq:iuvd_feature}.
\begin{equation} \label{eq:iuvd_feature}
    \mathcal{F}_{iuvd}(\tilde{P}) = [sdf(d), \tilde{N}^{s}(i,u,v), g(\tilde{N}_{vis}^{c}(i,u,v,d))],
\end{equation}
where $\tilde{N}_{vis}^c(i,u,v,d) = N_{vis}^c(x_p)$ is the visible \textit{IUV cloth normal} shown in Fig. \ref{fig:intro_overview}.

Note that the correspondence between $P$ and $P^b$ is not always unique, so the linear transformation $\mathcal{L}$ is not invertible. Figure \ref{fig:method_equivalence_analysis} (b) shows a non-unique case when the nearby surface is concave. To build a {unique dense correspondence} from XYZ space to IUVD space, we introduce a set of \textit{convex assumptions} on $P$ and $P^b$. When these assumptions are satisfied, $P$ is defined as a \textit{valid} query point.
\begin{assump} \label{assump:1}
    The neighbourhood of $P^b$ on ${S^b}$ is convex.
\end{assump}
\begin{assump} \label{assump:2}
    The query point $P$ is not so far from ${S^b}$.
\end{assump}

\textit{Discussion on Assumption \ref{assump:1}.}
Each part of the human body, except the head, hands and feet, can be considered as a rigid object. Following the physical hypothesis that the spherical surface is the most stable shape, the regular surface of each rigid body part should be convex when viewed from the outside. As a result, for query points outside the SMPL surface, the source point is unique. But for inner points, there may exist multiple source points. So the range of the D-axis should be limited to a minimum value, $d>D_{min}$.

\textit{Discussion on Assumption \ref{assump:2}.} 
When $sdf(d)$ increases, the query point $P$ will be generated far from the SMPL surface. But the precision of $P$ is limited by the precision of $N^b(P^b)$, which is restricted by the triangle mesh. So the range of the D-axis should also be limited to a maximum value, $d<D_{max}$. In addition, when considering different body parts, there will be many alternatives of the source point for a single query point. Therefore, the IUVD occupancy function should be visualized in separate UVD spaces for different body parts.

Consequently, the convex assumptions are satisfied when $d \in (D_{min}, D_{max})$ in each UVD space. For simplicity, we assume that all of the query points near the SMPL surface, except hands and feet, are {valid}. This is proved to be acceptable in our experiments. For each {valid} query point $\tilde{P} \in \Psi$ and $P \in \Phi$, the IUVD features $\mathcal{F}_{iuvd}(\tilde{P})$ are equal to the XYZ features $\mathcal{F}_{xyz}(P)$ as a result of the convex assumptions. This equivalence property ensures the reconstruction accuracy without retraining the neural networks.

\begin{figure*}[!tp]
    \centering
    \includegraphics[width=0.96\linewidth]{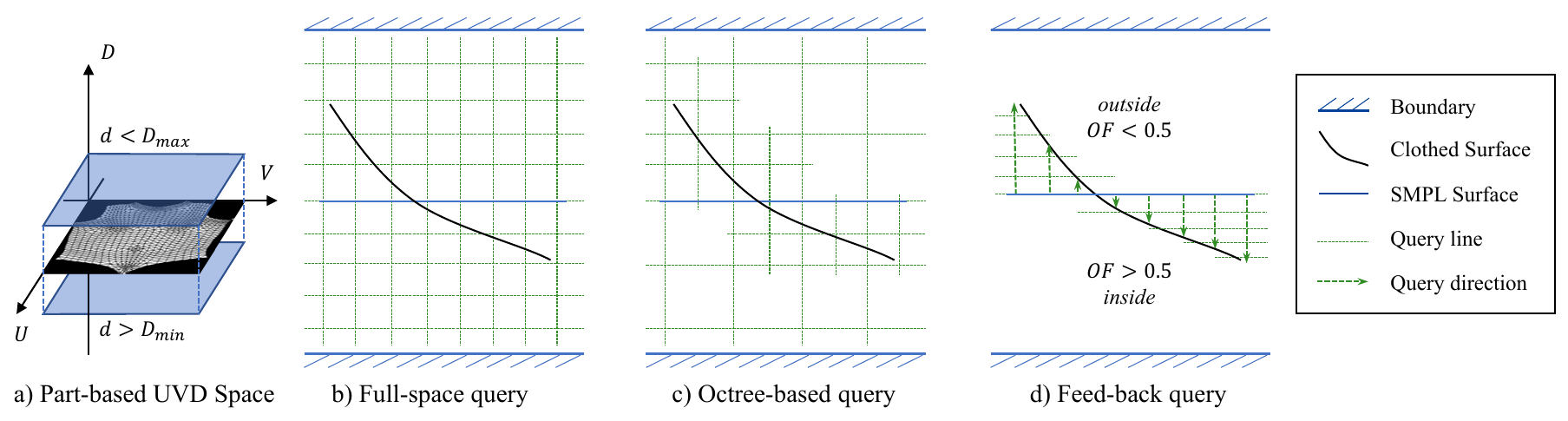}
    \caption{Illustration of the three query methods in IUVD space, i.e. the collection of part-based UVD spaces. In figure b), c), and d), the \textcolor{blue}{blue} middle lines denote the SMPL surface. The black curves denote the clothed human surface, where the occupancy value $\text{OF}=0.5$. The intersections of the \textcolor{green}{green} dotted lines, which denotes the query lines, and the \textcolor{green}{green} dotted lines with arrow, which denotes the query directions, are generated as query points.}
    \label{fig:method_query_methods}
\end{figure*}

\subsection{IUVD Feedback Query} \label{sec:iuvd_query}
To localize the 3D human surface from the implicit IUVD occupancy function, different query-and-infer algorithms can be used. In this section, we present three algorithms, including full-space query, octree-based query, and a novel feedback query method to accelerate surface localization.

\begin{algorithm}[!tp]
	\renewcommand{\algorithmicrequire}{\textbf{Input:}}
	\renewcommand{\algorithmicensure}{\textbf{Output:}}
	\caption{IUVD Feedback query}
	\label{alg:iter_query}
	\begin{algorithmic}[1]
	
	    \REQUIRE SMPL UV maps $\tilde{M}$, IUVD local body features $\mathcal{F}_{iuvd}(\tilde{P})$ for each query point $\tilde{P}(i,u,v,d)$
	    
		\STATE \textbf{Initialization:}
		\FOR{$\tilde{P}^b(i,u,v) \in \tilde{M}$}
		\STATE $\tilde{P} \gets (i,u,v,0)$
		\STATE $f_{\theta}(\tilde{P}) \gets \text{MLP}(\mathcal{F}_{iuvd}(\tilde{P}))$
		\IF{$f_{\theta}(\tilde{P})>0.5$}
		\STATE $\delta(i,u,v) \gets +1$
		\STATE $f_{\theta}(\tilde{P}_{inner}) \gets \text{OF}_{max}$
		\ELSE
		\STATE $\delta(i,u,v) \gets -1$
		\STATE $f_{\theta}(\tilde{P}_{outer}) \gets \text{OF}_{min}$
		\ENDIF
		\ENDFOR
		
		\STATE \textbf{Iteration:}
		\FOR{$\tilde{P}^b(i,u,v) \in \tilde{M}$}
		\STATE $d \gets 0$
		\REPEAT
		\STATE $f_{\theta}(\tilde{P})^{pre} \gets f_{\theta}(\tilde{P})$
		\STATE $d \gets d + \delta(i,u,v)$
		\STATE $\tilde{P} \gets (i,u,v,d)$
		\STATE $f_{\theta}(\tilde{P}) \gets \text{MLP}(\mathcal{F}_{iuvd}(\tilde{P}))$
		\UNTIL{$(f_{\theta}(\tilde{P}) - 0.5 ) \cdot (f_{\theta}(\tilde{P})^{pre} - 0.5) < 0$ \\ or $d > D_{max}$ or $d < D_{min}$}
		\STATE $f_{\theta}(\tilde{P}_{remains}) \gets \text{sign}(0.5-f_{\theta}(\tilde{P})^{pre}) \cdot \text{OF}_{max}$
		\ENDFOR
		
		\ENSURE IUVD occupancy values $\{f_{\theta}(\tilde{P})\} \in \mathbb{R}^{I\times U\times V\times D}$
		
	\end{algorithmic}  
\end{algorithm}

\textit{1) Full-space query \& Octree-based query.}
A simple idea of the query-and-infer process is to evaluate the IUVD occupancy function in the whole IUVD space, so called the \textit{full-space query}. Based on the voxel grid of each UVD space, all of the voxels $\tilde{P}$ will be visited and the corresponding IUVD local body features are fed into a MLP to predict the occupancy value $f_{\theta}(\tilde{P}) = \text{MLP}(\mathcal{F}_{iuvd}(\tilde{P}))$. For acceleration, the \textit{octree-based surface localization} algorithm \cite{li2020monocular} is adopted to subdivide the voxels near the human surface iteratively.

However, experiments show that the above two query methods always fail to generate a reasonable result in IUVD space. The reconstructed surface is always discontinuous, especially when the body parts are closely interacted. There are two possible reasons for this. First, the sampling points far away from the body surface, whose occupancy values are not accurate, as indicated by the convex assumption \ref{assump:2}, will disturb the result of marching cubes. Second, if the sampling point is far away from the body surface, it may be inside the other body parts, resulting in multiple discontinuous surfaces.

\textit{2) Continuity assumptions.}
To solve the discontinuity problem, we introduce two assumptions as follows.
\begin{assump} \label{assump:3}
    The clothed human surface is a single, continuous layer of mesh between the skin and the clothing.
\end{assump}
\begin{assump} \label{assump:4}
    The IUVD occupancy function is continuous and locally monotonic along the D-axis.
\end{assump}

The two assumptions ensure the completeness of the results and are easy to satisfy. Under the continuity assumptions, we formulate the implicit surface reconstruction as a locally convex optimization problem, the goal of which is equivalent to finding the optimal $d$ value at each $(i,u,v)$ coordinate.

\textit{3) Feedback query.}
Based on the continuity assumptions, we introduce a feedback mechanism into the query-and-infer process, where the IUVD occupancy function is evaluated in a directional and iterative manner, as shown in Algorithm \ref{alg:iter_query}.

In initialization, the occupancy values of the points lying on the SMPL UV maps, $\tilde{P}^b(i,u,v) \in \tilde{M}$, are inferred using Eq. \eqref{eq:occupancy_infer}. These values generate the \textit{query directions}, $\delta(i,u,v)$, parallel to the D-axis. Depending on the query directions, the inner/outer points $\tilde{P}_{inner}, \tilde{P}_{outer}$ at $(i,u,v)$ are set to a maximum/minimum value without inference. This reduces the number of query points by half.

In each iteration, the $d$ value at $(i,u,v)$ is updated along $\delta(i,u,v)$. The current batch of query points is then generated by Eq. \eqref{eq:pb_to_p}. And their occupancy values are inferred using Eq. \eqref{eq:occupancy_infer}. The iteration at $(i,u,v)$ will stop when the current query point and the previous one are on the opposite side of the clothed human surface, which means the single layer surface is localized. The remaining query points $\tilde{P}_{remains}$ at $(i,u,v)$ are set to a maximum/minimum value according to their relationship to the surface. Meanwhile, if the D-axis boundary is reached, the iteration will also be terminated.

Note that the complexity of a query algorithm is closely related to the number of query points. In Fig. \ref{fig:method_query_methods}, the query points generated by the three different query methods are denoted as the intersections of query lines and query directions. With similar resolution, the feedback query method produces much fewer but more reasonable query points, compared to the full-space and octree-based query methods.

\begin{figure}[!tp]
    \centering
    \includegraphics[width=0.99\linewidth]{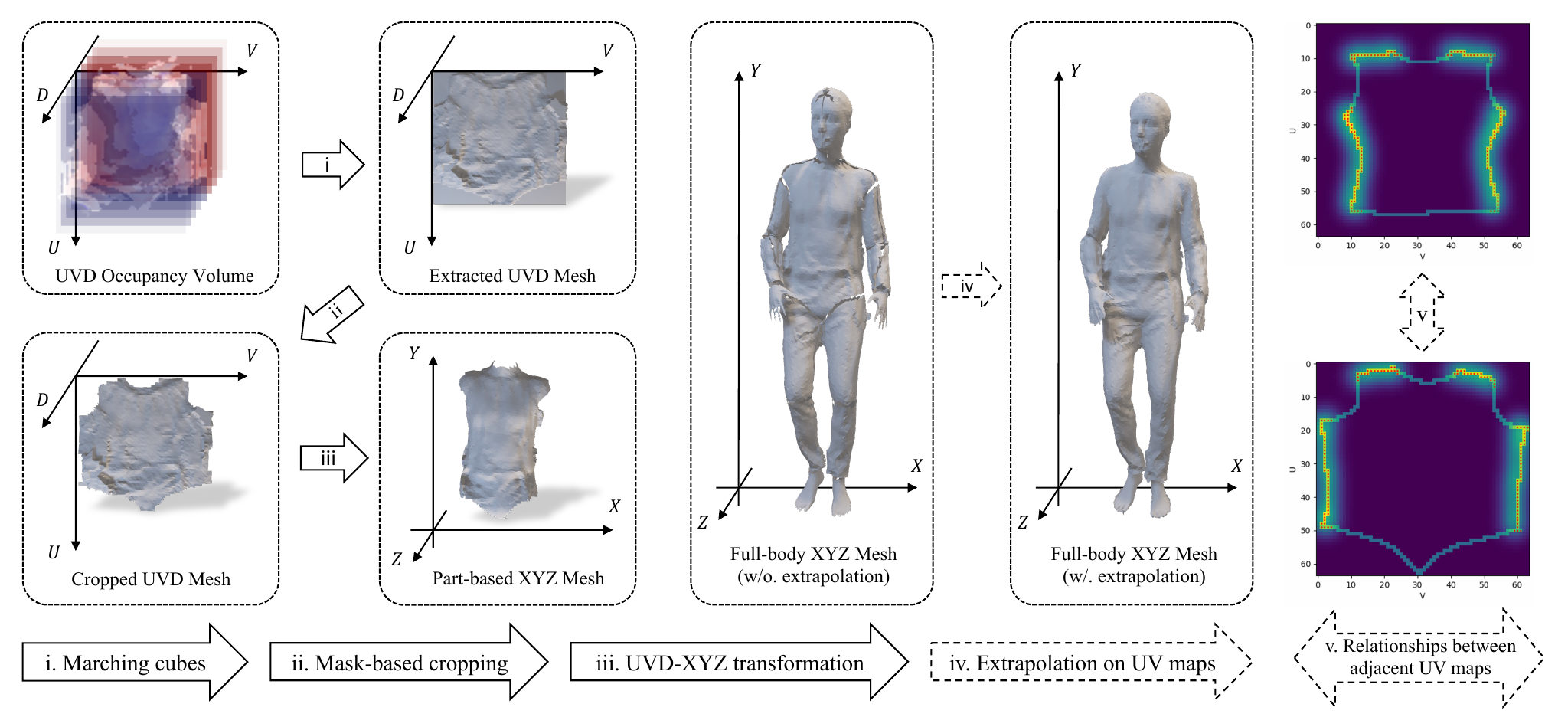} 
    \vspace{-5mm}
    \caption{Illustration of the mesh transformation steps from IUVD space to XYZ space. Note that the steps (iv and v) with a dotted arrow box indicate the effect of offline pre-processing and do not affect the runtime.}
    \label{fig:exp_mesh_transform}
    \vspace{-2mm}
\end{figure}

\subsection{IUVD to XYZ Mesh Transformation} \label{sec:iuvd_mesh}

To obtain a visually watertight result in XYZ space, we design a realtime approach that combines offline dilation and online erosion steps, instead of using the time-consuming Poisson surface reconstruction \cite{kazhdan2006poisson}.

In the offline preprocessing, the SMPL UV maps are dilated with a square structuring element and then filled using bilinear extrapolation, which are designed to fill the marginal gaps between adjacent body parts.

During the online reconstruction, we apply the marching cubes algorithm \cite{lorensen1987marching} in each UVD space to separately extract the triangle meshes of different body parts from the IUVD occupancy function, as shown in Fig. \ref{fig:exp_mesh_transform} (i). Note that the marching cubes algorithm takes only 8 ms in IUVD space, which is much faster than in XYZ space. Then, a UV mask-based cropping step is applied, as shown in Fig. \ref{fig:exp_mesh_transform} (ii), to erode the undesired edges of the part-based meshes in IUVD space. With GPU-based parallel acceleration, the additional cropping step costs less than 10 ms. Finally, as shown in Fig. \ref{fig:exp_mesh_transform} (iii), the part-based meshes are transformed to XYZ space by applying a linear transformation as Eq. \eqref{eq:pb_to_p} to each vertex. Note that the offline preprocessing does not affect the online running time, but makes the final result more complete like a whole body, as shown in Fig. \ref{fig:exp_mesh_transform} (iv).

In addition, it is theoretically possible to obtain a topologically watertight result based on the relationships between adjacent UV maps. For example, Fig. 7 (v) highlights the related pixels between two adjacent UV maps. Based on such relationships, we can obtain a whole mesh that combines different body parts by connecting the related vertices or triangles of adjacent UVD part meshes.

\section{Experiments} \label{sec:experiments}

\begin{table*}[!tp]
\centering
\caption{Running time (in milliseconds) of ICON \cite{xiu2022icon} with different representations at matching resolutions. The \textit{SDF calculation} and \textit{MLP regression} are two main steps in the \textit{query-and-infer} process. The \textit{surface extraction} is to obtain the human mesh, including but not limited to marching cubes. Note that the pre-processing steps (e.g. segmentation \cite{gatis2022rembg}, HPS estimation \cite{zhang2021pymaf}) and the cloth-normal refinement are not included in this table since they are optional and replaceable in implicit reconstruction.}
\resizebox{0.99\linewidth}{!}{
    \begin{tabular}{cccccc}
    \toprule
    \diagbox[dir=NW]{Main steps}{Representations} & \makecell{XYZ-Full \\$(257^3)$} & \makecell{XYZ-Octree \cite{xiu2022icon} \\$(257^3)$} & \makecell{IUVD-Full \\ $(24\times 64^2 \times 21)$} & \makecell{IUVD-Octree \\ $(24\times 64^2 \times 21)$} & \makecell{IUVD-Feedback (Ours) \\ $(24\times 64^2 \times 21)$} \\
    \midrule

    {SDF calculation}    & 3870 & 52 & 2  & 4  & \textbf{2} \\
    {MLP regression}     & 957  & 27 & 26 & 14 & \textbf{7} \\
    {Surface extraction} & 33   & 25 & 18 & 18 & \textbf{18} \\
    \midrule

    Query-and-infer & 5.1k (5.0k$\sim$5.3k) & 98 (74$\sim$155) & 36 (33$\sim$38) & 28 (25$\sim$33) & \textbf{27} (21$\sim$34) \\
    Total (single thread) & 5.3k (5.2k$\sim$5.5k) & 257 (238$\sim$310) & 183 (178$\sim$193) & 176 (173$\sim$181) & \textbf{175} (168$\sim$189) \\
    \bottomrule
    \end{tabular}
}
\label{tab:speed_comparison_1}
\end{table*}

\begin{table*}[!tp]
\centering
\caption{Comparison on the number of query points and the marching cubes complexity of ICON \cite{xiu2022icon} with different representations. The resolutions of the XYZ and IUVD space are denoted as $N=257^3$ and $M=24\times 64^2 \times 21$ correspondingly. Note that $N > M$.}
\resizebox{0.90\linewidth}{!}{
    \begin{tabular}{cccccc}
    \toprule
    Representations & XYZ-Full & XYZ-Octree \cite{xiu2022icon} & IUVD-Full & IUVD-Octree & IUVD-Feedback (Ours) \\
    \midrule
    Number of query points & $1.6 \times 10^7$ & $1.2 \times 10^5$ & $4.2 \times 10^5$ & $2.2 \times 10^5$ & \textbf{$5.2 \times 10^4$} \\
    Marching cubes complexity & $O(N)$ & $O(N)$ & \textbf{$O(M)$} & \textbf{$O(M)$} & \textbf{$O(M)$} \\
    \bottomrule
    \end{tabular}
}   
\label{tab:speed_comparison_2}
\end{table*}

\subsection{Settings} \label{sec:settings}

\textit{1) Dataset and rendering.}
The THuman2.0 dataset \cite{yu2021function4d} is used for training and quantitative evaluation. It is a public dataset with 526 high-quality textured scans of clothed humans and fitted SMPL \cite{loper2015smpl}, SMPL-X \cite{pavlakos2019expressive} models. The first 500 scans are used for training. Another 26 scans are used for evaluation.

To obtain the image data, we render the scans of THuman2.0 dataset with a weak perspective camera as \cite{xiu2022icon}. Especially, the camera viewpoints consist of 12 horizontal and 3 elevation angles. It has been proved that the variation of elevation angles improves the model accuracy with the same amount of data.

\textit{2) Training and evaluation.}
To evaluate the speed and accuracy improvements brought by the proposed IUVD representations, we take PIFu \cite{saito2019pifu}, PaMIR \cite{zheng2021pamir}, and ICON \cite{xiu2022icon} as baselines. As for PIFu and PaMIR, the pretrained models are used. For ICON, we re-train the neural networks including the NormalNet, HGFilter and MLP with the THuman2.0 dataset for 20 epochs on a single NVIDIA GTX 3090 GPU. 

\begin{table}[!tp]
\centering
\caption{Accuracy comparison of PIFu \cite{saito2019pifu}, PaMIR \cite{zheng2021pamir}, and ICON \cite{xiu2022icon} with XYZ-Octree or IUVD-Feedback representations on THuman2.0. We also compare with other state-of-the-art methods including FOF \cite{feng2022fof}, IntegratedPIFu \cite{chan2022integratedpifu}, and ECON \cite{xiu2023econ}.}
\label{tab:accuracy_comparison}
\resizebox{0.95\linewidth}{!}{
    \begin{tabular}{lccc}
    \toprule
    Model (Representation) & P2S$\downarrow$ & Chamfer$\downarrow$ & Normal$\downarrow$ \\
    \midrule
    PIFu (XYZ-Octree) \cite{saito2019pifu} & 2.824 & 3.245 & 0.139 \\
    PIFu (\textbf{IUVD-Feedback}) & \textbf{1.561} & \textbf{1.645} & \textbf{0.116} \\
    \midrule
    PaMIR (XYZ-Octree) \cite{zheng2021pamir} & 1.304 & 1.941 & 0.105 \\
    PaMIR (\textbf{IUVD-Feedback}) & \textbf{1.059} & \textbf{1.224} & \textbf{0.080} \\
    \midrule
    ICON (XYZ-Octree) \cite{xiu2022icon} & \textbf{0.832} & 1.114 & 0.072 \\
    ICON (\textbf{IUVD-Feedback}) & 0.925 & \textbf{1.006} & \textbf{0.072} \\
    \midrule
    ICON-refine (XYZ-Octree) \cite{xiu2022icon} & \textbf{0.798} & 1.082 & 0.057 \\
    ICON-refine (\textbf{IUVD-Feedback}) & 0.822 & \textbf{0.906} & \textbf{0.057} \\
    \midrule
    FOF (w/o. SMPL) \cite{feng2022fof} & 3.325 & 3.184 & 0.128 \\
    IntegratedPIFu \cite{chan2022integratedpifu} & 1.215 & 1.282 & 0.070 \\
    ECON \cite{xiu2023econ} & \textbf{1.097} & \textbf{1.081} & \textbf{0.065} \\
    \bottomrule
    \end{tabular}
}
\vspace{-1.5mm}
\end{table}%

\begin{table}[!tp]
\centering
\caption{The average reconstruction error of the scans in THuman2.0 dataset using IUVD-Full and IUVD-Feedback representations.}
\resizebox{0.95\linewidth}{!}{
    \begin{tabular}{lccc}
    \toprule
    Representation & P2S$\downarrow$ & Chamfer$\downarrow$ & Normal$\downarrow$ \\
    \midrule
    IUVD-Full & \textbf{0.095} & \textbf{0.075} & 0.032 \\
    IUVD-Feedback & \underline{0.159} & 0.291 & \underline{0.025} \\
    IUVD-Feedback (Poisson) & 0.172 & \underline{0.273} & \textbf{0.023} \\
    \bottomrule
    \end{tabular}
}
\vspace{-3.5mm}
\label{tab:accuracy_iuvd_gt}
\end{table}%

For speed evaluation, we take ICON as the baseline model and conduct the query-and-infer process using five different representations. Two of them are in XYZ space, including \textit{XYZ-Full} (using full-space query) and \textit{XYZ-Octree} (using octree-based query with three levels). Three of them are in IUVD space, including \textit{IUVD-Full} (using full-space query) and \textit{IUVD-Octree} (using octree-based query with two levels), and \textit{IUVD-Feedback} (using the proposed feedback query). The resolutions of the XYZ space and the IUVD space are set to $257 \times 257 \times 257$ and $24 \times 64 \times 64 \times 21$ to ensure a similar precision. Correspondingly, the scale factor $\alpha$ is set to $1/128$. $D_{max}=10$, and $D_{min}=-10$. For acceleration, we use the GPU-based marching cubes function of NVIDIA Kaolin \cite{jatavallabhula2019kaolina} library to extract surface. For comparison, we report the detailed running time of ICON using the five representations in Table \ref{tab:speed_comparison_1}, where the input image is shown in Fig. \ref{fig:intro_overview}. The test is repeated for 30 times to avoid random errors. We also compare the number of query points and the marching cubes complexity between different representations in Table \ref{tab:speed_comparison_2}.

\begin{figure}[!tp]
    \centering
    \includegraphics[width=0.99\linewidth]{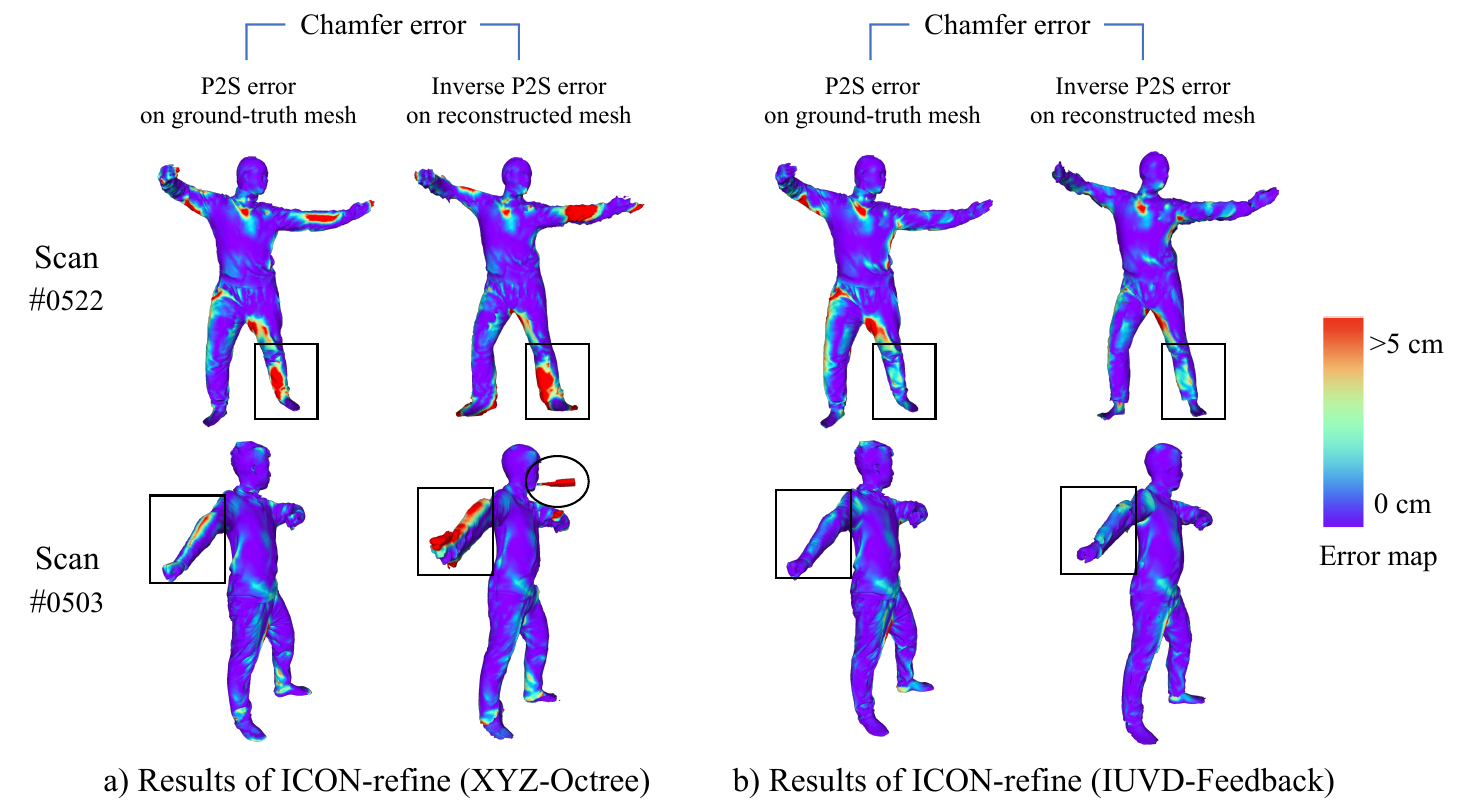}
    \caption{Visualization of the P2S error on the ground-truth mesh and the inverse P2S error on the reconstructed mesh, whose average value is the Chamfer error. The mis-reconstructed limbs and the stitched artifacts marked by rectangles and circles are called the ``redundant reconstruction artifacts''.}
    \label{fig:exp_error_maps}
\end{figure}

\begin{figure*}[htbp]
    \centering
    \includegraphics[width=0.99\linewidth]{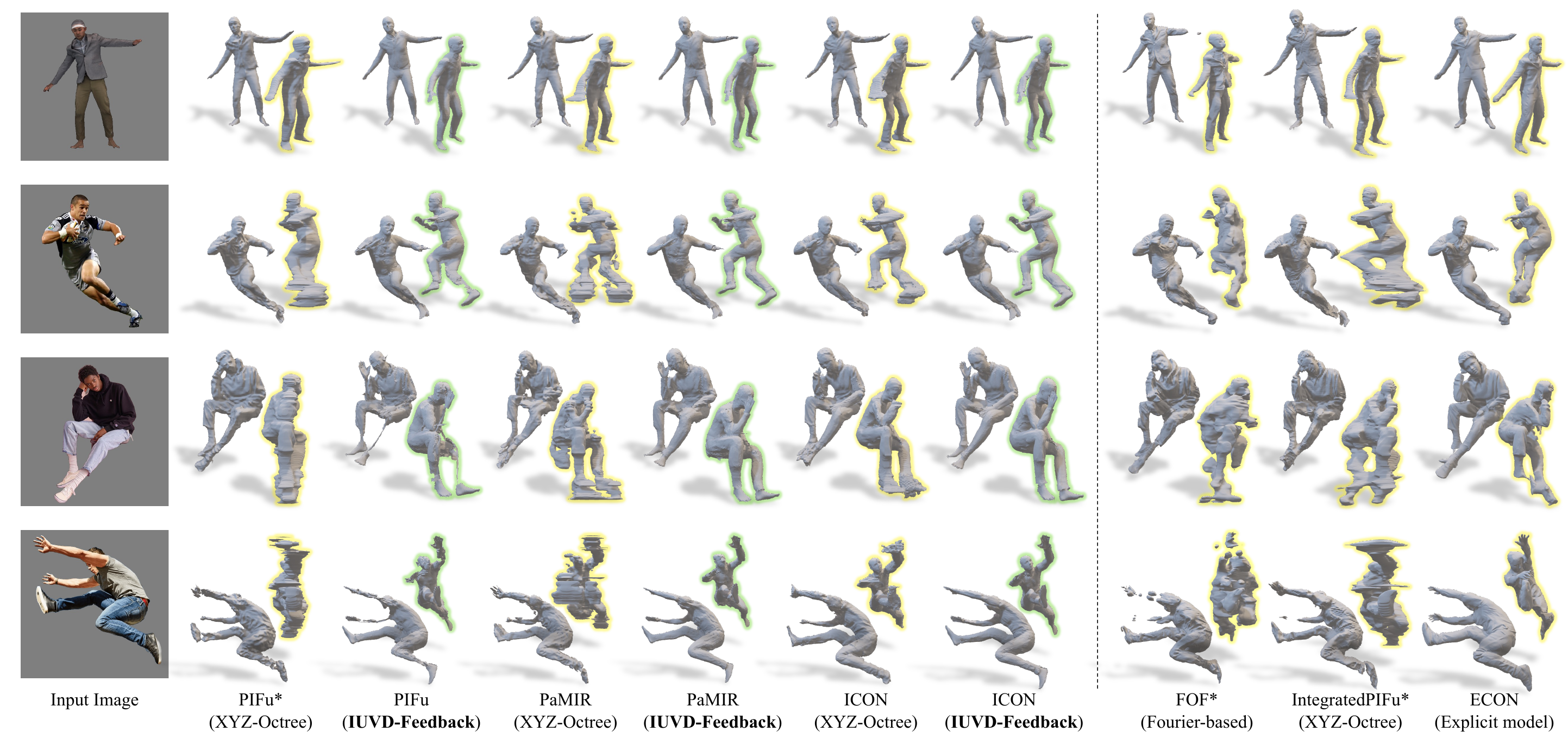}
    \vspace{-2mm}
    \caption{Qualitative comparison on in-the-wild images with various human poses. The main results are obtained by PIFu \cite{saito2019pifu}, PaMIR \cite{zheng2021pamir} and ICON \cite{xiu2022icon} (without cloth-normal refinement) models with the XYZ-Octree \cite{li2020monocular} and the proposed IUVD-Feedback representations. Results from other state-of-the-art methods including FOF \cite{feng2022fof}, IntegratedPIFu \cite{chan2022integratedpifu}, and ECON \cite{xiu2023econ} are also compared. The results highlighted with \hlyellow{yellow} and \hlgreen{green} edges are observed from side views, which show the robustness of the proposed IUVD-Feedback representation. The $*$ denotes that the method does not use parametric body model.}
    \label{fig:exp_qualitative_comparison}
    \vspace{-2mm}
\end{figure*}

For accuracy evaluation, we take PIFu \cite{saito2019pifu}, PaMIR \cite{zheng2021pamir}, and ICON \cite{xiu2022icon} as the baseline models, and compare the reconstruction accuracy of these models using XYZ-Octree and IUVD-Feedback representations. We also compare the proposed method with well-known and state-of-the-art methods including FOF \cite{feng2022fof}, IntegratedPIFu \cite{chan2022integratedpifu}, and ECON \cite{xiu2023econ}. The evaluation metrics include the {point-to-surface distance} (\textit{P2S}) and the {chamfer distance} (\textit{Chamfer}) between the predicted 3D meshes and ground-truth scans, as well as the {L2 distance between the rendered normal images} (\textit{Normal}). Note that the P2S error is computed by sampling points on ground-truth scan and then calculating the average value of their distances to the nearest points on the predicted mesh. The Chamfer error is computed by averaging the P2S error and the inverse P2S error that sampling points on the predicted mesh. When using the IUVD-Feedback representation, the SMPL mesh of hands and feet is preserved to obtain more robust results. We report the results of ICON using the offline cloth-normal refinement, denoted as {ICON-refine}. The refinement step, as used in \cite{xiu2022icon}, defines an iterative local affine transformation for the vertices of the predicted mesh to optimize its rendered normal maps based on the estimated cloth normal maps. The quantitative evaluation results on THuman2.0 dataset are shown in Table \ref{tab:accuracy_comparison}. For qualitative comparison, Fig. \ref{fig:exp_qualitative_comparison} shows the visualization of reconstruction results for in-the-wild images with various human poses, and Fig. \ref{fig:exp_clothing_details} compares the clothing details between the XYZ-Octree and IUVD-Feedback representations.

\subsection{Speed Evaluation} \label{sec:speed_comparison}

Considering the barrel effect, we mainly compare the three most time-consuming steps of ICON, including SDF calculation, MLP regression, and surface extraction.

\textit{1) SDF calculation.} 
As shown in Table \ref{tab:speed_comparison_1}, the SDF calculation is almost the most time-consuming step when using XYZ-Full and XYZ-Octree representations. However, by replacing the P2S optimization with a linear transformation (see Fig. \ref{fig:method_features_composition}) in IUVD space, the time of this step is significantly reduced.

\textit{2) MLP regression.}
Note that the MLP regression time is approximately in proportion to the number of query points. Table \ref{tab:speed_comparison_2} shows that the IUVD-Feedback representation reduces the number of query points by $87.7\%$ than IUVD-Full and by $54.8\%$ than XYZ-Octree. And there is almost no decrease in the accuracy of ICON, as shown in Table \ref{tab:accuracy_comparison}. This proves that the IUVD-Feedback representation successfully reduces the redundancy in the implicit query-and-infer process.

\begin{figure}[!tp]
    \centering
    \includegraphics[width=0.99\linewidth]{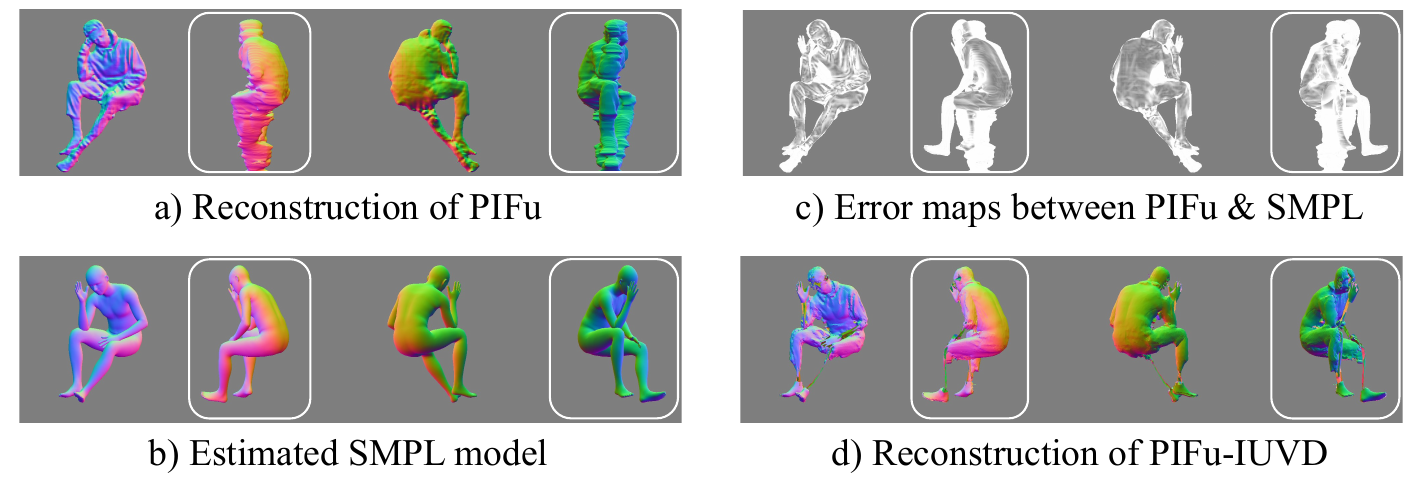}
    \caption{Analysis on the failure case of PIFu-IUVD. When the original prediction of PIFu is not aligned with the fitted SMPL model, the result of IUVD-Feedback will be incomplete, thus generating super thin limbs.}
    \label{fig:exp_pifu_iuvd_analysis}
    \vspace{-4mm}
\end{figure}

\textit{3) Surface extraction.}
To obtain an explicit mesh from the implicit function, the marching cubes algorithm \cite{lorensen1987marching} is always required, the time consumption of which is related to the surface geometry and the space resolution \cite{huang2013marching}. Moreover, in IUVD space, we need additional online erosion and linear transformation for reconstructing a full human mesh (see Sec. \ref{sec:iuvd_mesh}). Totally, the surface extraction time of IUVD-Feedback representation is $72\%$ than that of the XYZ-Octree representation. When comparing the time cost of marching cubes, it is shown that the algorithm complexity in IUVD space is $12.2\%$ than that in XYZ space.

\textit{4) Overall comparison.}
In summary, the query-and-infer process using IUVD-Feedback representation is over three times faster than using XYZ-Octree representation, which helps to reduce the overall runtime by about $32\%$ on average. The results prove that the proposed IUVD-Feedback representation is efficient for 3D clothed human surface reconstruction.

\subsection{Accuracy Evaluation} \label{sec:accuracy_robustness}

\textit{1) Quantitative comparison on representations.}
Table \ref{tab:accuracy_comparison} shows that the IUVD-Feedback representation improves the accuracy of PIFu and PaMIR in all metrics. This is because that PIFu carries no prior of SMPL, and PaMIR lacks the out-of-distribution pose prior of SMPL, which can be complemented by the IUVD representation. As for ICON, the results of the two representations have similar accuracy on average, since the SMPL body prior has been utilized by the local body features \cite{xiu2022icon}. From Table \ref{tab:accuracy_comparison}, we notice that the P2S errors of ICON and ICON-refine perform in opposite to the Chamfer errors when changing the representations. To find out the reasons, we visualize the P2S error on both the ground-truth scan and the reconstructed surface in Fig. \ref{fig:exp_error_maps}. Since the P2S error is defined on the ground-truth mesh, it alleviates the reconstruction error in global shapes, e.g. the mis-reconstructed limbs (marked by black rectangles) and the `stitched artifacts' (circled in black, whcih is possibly caused by self-occlusion) that cannot be cleaned by post-processing.

\begin{figure}[!tp]
    \centering
    \includegraphics[width=0.85\linewidth]{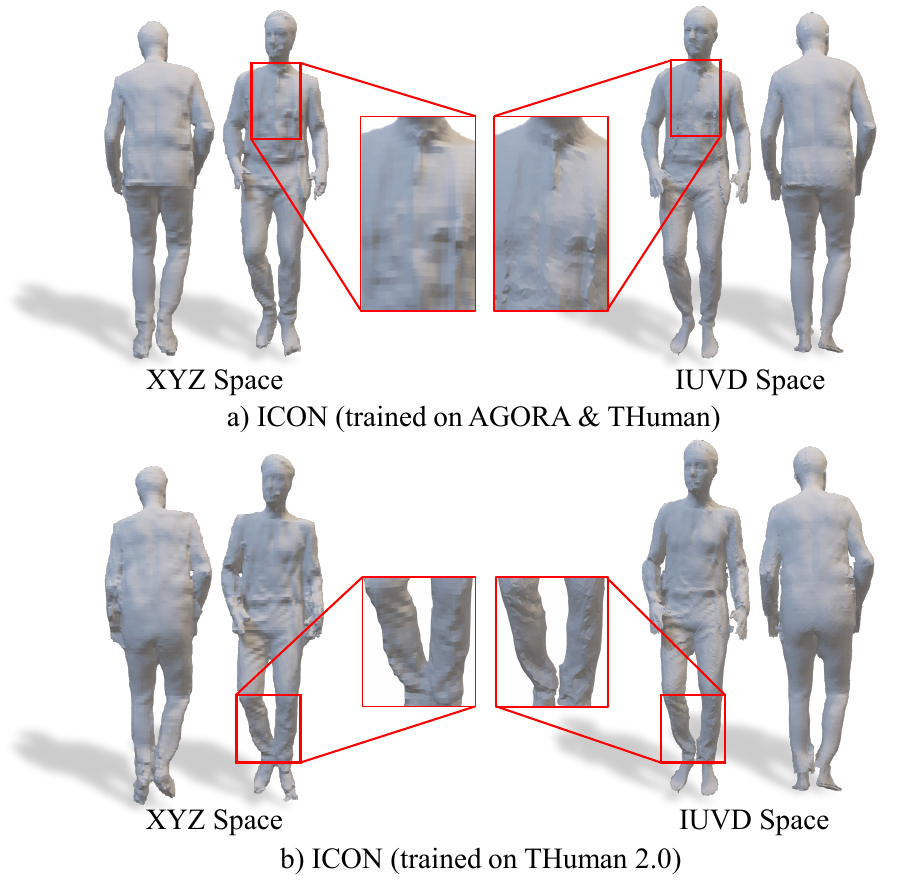}
    \caption{Comparison on clothing details reconstructed by ICON in XYZ and IUVD spaces. The feature equivalence is not effected by the training process.}
    \label{fig:exp_clothing_details}
\end{figure}

\textit{2) Qualitative comparison on representations.}
As shown in Fig. \ref{fig:exp_qualitative_comparison}, the IUVD-Feedback representation improves the robustness of most results, compared to the XYZ-Octree representation. As for PIFu, the IUVD-Feedback makes the side view of the results more recognizable. But when the original prediction of PIFu is not properly fitted with the SMPL model, the misaligned parts can not be reconstructed in IUVD space, thus generating super thin limbs. We illustrate such failure case in Fig. \ref{fig:exp_pifu_iuvd_analysis}. As for PaMIR, the mis-estimated out-of-body parts are eliminated by the IUVD-Feedback, thus making the results look more cleaner. As for ICON, the IUVD-Feedback produces more reasonable results especially for the limbs of human, where the hands and feet mesh is replaced by the corresponding parts of SMPL mesh. To sum up, the IUVD-Feedback representation makes the reconstruction are more humanlike than the previous results.

\textit{3) Comparison on clothing details.}
To prove the equivalence of the local body features in XYZ and IUVD space (see Sec. \ref{sec:iuvd_feature}), we compare the clothing details reconstructed by ICON using different representations of the two spaces. As shown in Fig. \ref{fig:exp_clothing_details}, the results reconstructed by XYZ and IUVD representations share the same clothing shape, and the IUVD representation even enhances the geometric details. Meanwhile, a different ICON model is also used for comparison, which is trained on the AGORA \cite{patel2021agora} and THuman \cite{zheng2019deephuman} datasets by \cite{xiu2022icon}. It proves that the feature equivalence property is not influenced by the training process.

\begin{figure}[!tp]
    \centering
    \includegraphics[width=0.87\linewidth]{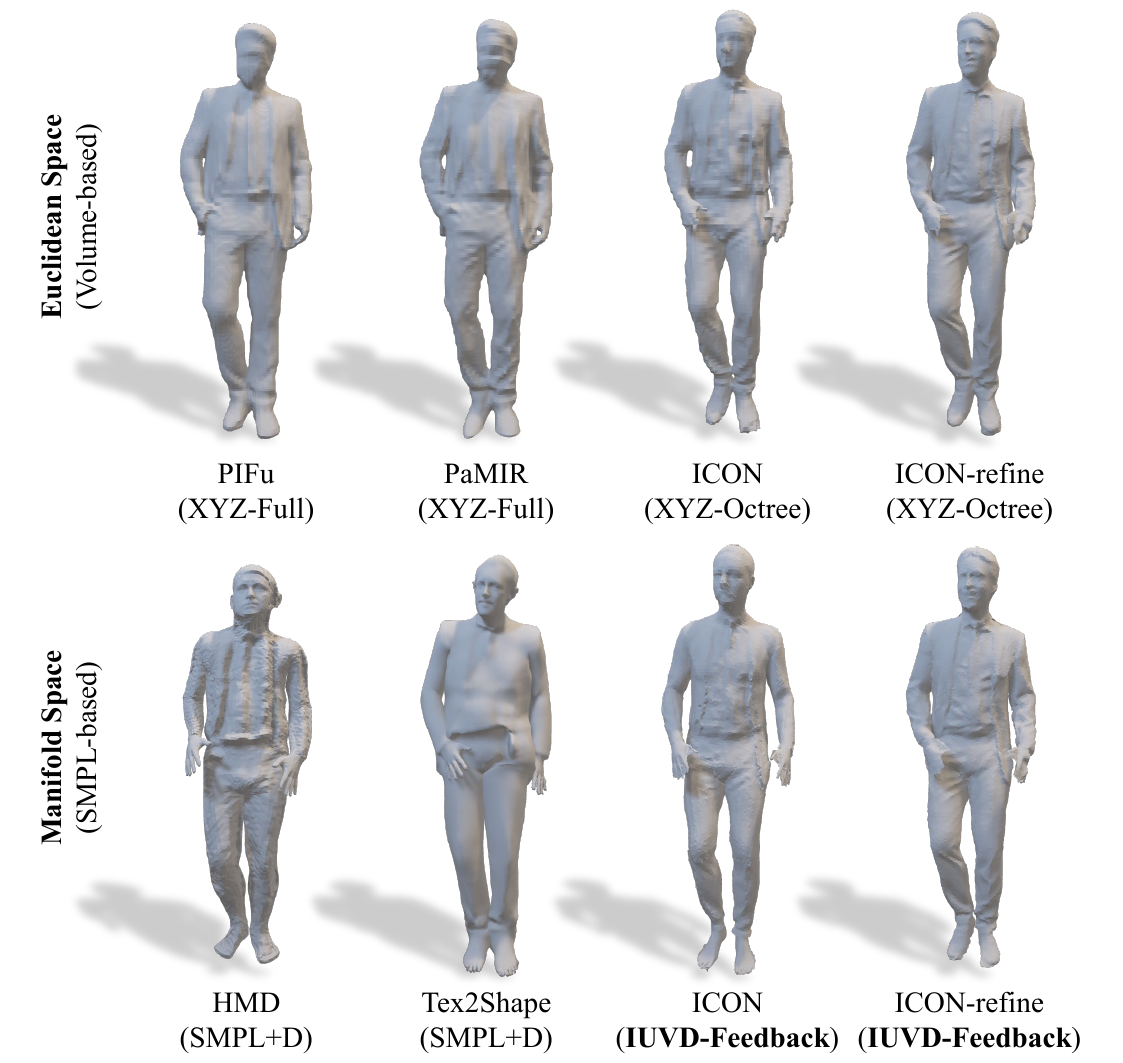}
    \caption{Comparison on 3D clothed human surface representations in the volume-based Euclidean (XYZ) space and the SMPL-based manifold space.}
    \label{fig:exp_comparison_others}
\end{figure}

\textit{4) Comparison with other state-of-the-art methods.}
We compare the proposed method with FOF \cite{feng2022fof}, IntegratedPIFu \cite{chan2022integratedpifu}, and ECON \cite{xiu2023econ} in Table \ref{tab:accuracy_comparison} and Fig. \ref{fig:exp_qualitative_comparison}. 
Firstly, we use the publicly available model of FOF \cite{feng2022fof}, which is also trained on THuman2.0 dataset but does not use the fitted SMPL models. For a fair comparison, we test FOF on images with variations in only horizontal viewpoints. Experimental results show that FOF lacks generalization ability to deal with unseen poses, although its running speed is over 30 fps. 
Secondly, we reimplement IntegratedPIFu \cite{chan2022integratedpifu} by revising its open-source code, and train it on THuman2.0 dataset with the same settings as described in Sec. \ref{sec:settings}. Given that IntegratedPIFu consists of a high-resolution integrator capable of perceiving more detailed features, the reconstructed details in the front views are better than other implicit methods. But it struggles to accurately reconstruct reasonable side-view body shapes. 
Thirdly, for ECON \cite{xiu2023econ}, whose performance relies heavily on the accuracy of the SMPL-X \cite{pavlakos2019expressive} model, we use the ground-truth SMPL-X model from THuman2.0 dataset in our quantitative evaluation. Table \ref{tab:accuracy_comparison} shows that ECON outperforms ICON with XYZ-Octree representation in terms of Chamfer error, but does not surpass the IUVD-Feedback representation. Qualitatively, ECON excels in recovering clothing details due to its iterative normal integration process. However, it sometimes fails to keep a reasonable body shape, particularly for limbs, where our proposed method performs better.

\subsection{Discussion} \label{sec:discussion}

\begin{figure}[!tp]
    \centering
    \includegraphics[width=0.99\linewidth]{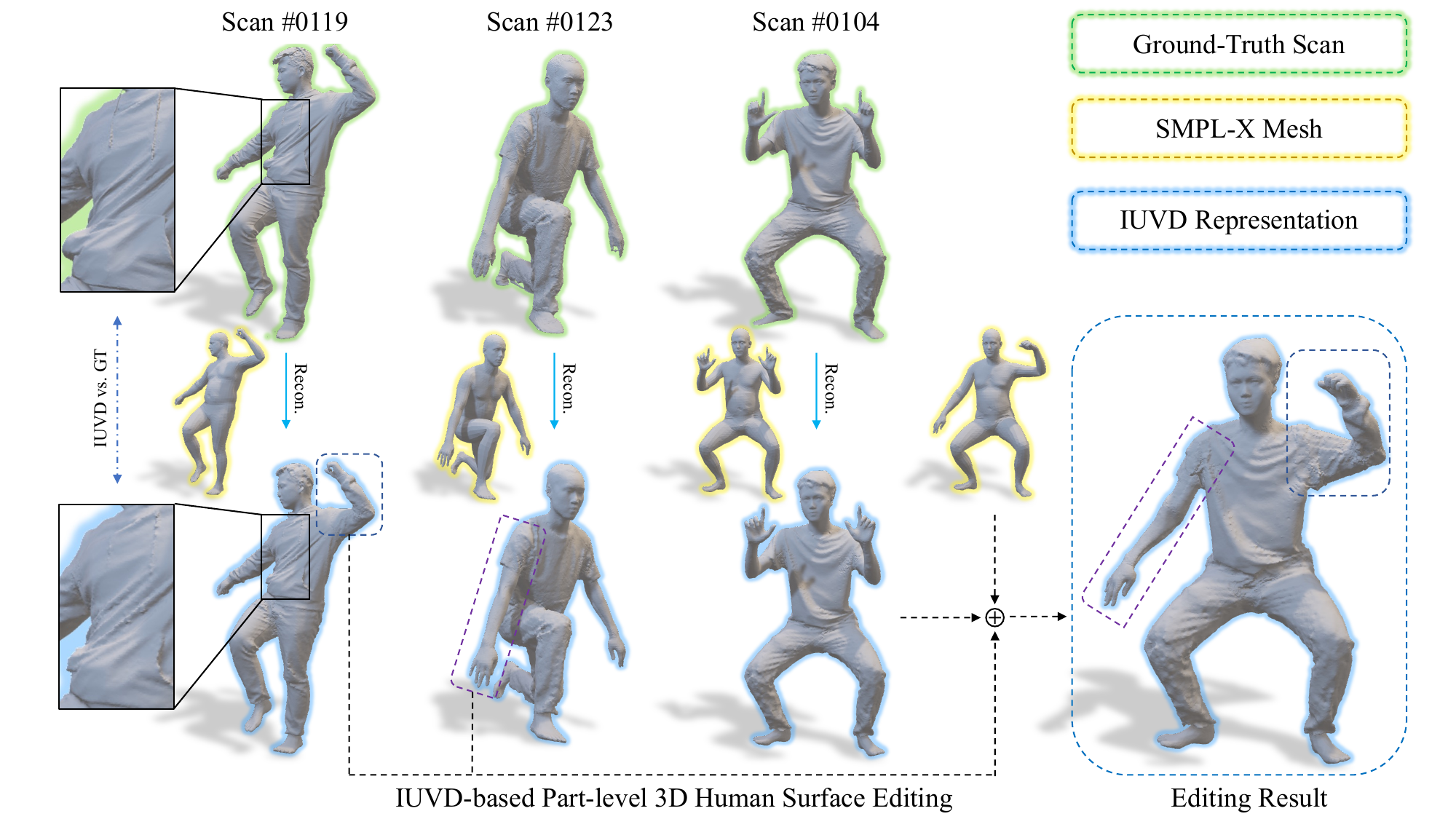}
    \caption{Reconstruction of the ground-truth scans in THuman2.0 dataset using the IUVD-Feedback representation. The results are then used for part-level 3D human surface editing application.}
    \label{fig:exp_part_editing_v2}
\end{figure}

\textit{1) Volume-based Euclidean space vs. SMPL-based manifold space.}
The 3D clothed human surface has been represented either by volume-based representations in Euclidean space, e.g. PIFu \cite{saito2019pifu} and PaMIR \cite{zheng2021pamir}, or SMPL-based surface deformation in manifold space, e.g. HMD \cite{zhu2019detailed} and Tex2Shape \cite{alldieck2019tex2shape}, as shown in Fig. \ref{fig:exp_comparison_others}. The proposed IUVD representation \textit{bridges the gap between the two spaces} by bringing the volume-based query-and-infer process into the SMPL-based manifold space. It combines the merits of both volume-based and SMPL-based approaches, including the pixel-aligned features, unlimited resolution, and the parametric body prior.

\textit{2) Upper limit of the IUVD representations.}
To evaluate the upper limit of the accuracy of the IUVD-based representations, we design an ideal experiment based on THuman2.0 dataset \cite{yu2021function4d}. Firstly, all of the ground-truth scans are transformed into SDF volumes, which are then used to replace the predicted IUVD occupancy values in the query-and-infer process. Secondly, we extract the part-based meshes and combine them in XYZ space as described in Sec. \ref{sec:iuvd_mesh}, thus obtaining the reconstructed human surfaces. Finally, we calculate the average reconstruction error using \textit{P2S}, \textit{Chamfer} and \textit{Normal} metrics, as shown in Table \ref{tab:accuracy_iuvd_gt}. In this experiment, the resolution of IUVD space is set to $24 \times 128 \times 128 \times 21$ and the scale factor $\alpha = 0.003$. The SMPL-X hands and feet meshes are preserved to prevent severe non-unique correspondence problem. In this experiment, we use the Poisson surface reconstruction \cite{kazhdan2006poisson} to smooth the reconstructed meshes for better visualization.

This ideal experiment draws two conclusions. Firstly, when comparing Table \ref{tab:accuracy_comparison} and Table \ref{tab:accuracy_iuvd_gt}, it is noticeable that the IUVD representations show great potential to achieve very high accuracy if the occupancy value can be accurately predicted, which may be achieved by properly designing and training the MLP network. This can also be seen in the visualization comparison, IUVD vs. GT, in Fig. \ref{fig:exp_part_editing_v2}. Secondly, compared to the full-space query, the feed-back query decreases the reconstruction accuracy in clothing details due to the continuity assumptions (see Sec. \ref{sec:iuvd_query}). But if these assumptions can be slacked, the upper limit of the IUVD representation will be raised from IUVD-Feedback to IUVD-Full as shown in Table \ref{tab:accuracy_iuvd_gt}, which deserves future research.

\textit{3) Application in part-level human surface editing.}
Based on the results of the above ideal experiment, we find that the semantic information of the IUVD representation can be utilized in generative applications. As shown in Fig. \ref{fig:exp_part_editing_v2}, by combining the part-based meshes of different scans, we can generate novel 3D scans with high-fidelity resolution. In the generation process, the SMPL-X model is used as an intermediate representation to preserve the body shape. So the generated result is naturally fitted with an SMPL-X model. As a result, it is a relatively inexpensive way to generate 3D human surface data, since the collection of high-fidelity 3D human scans is a rather expensive task.

\begin{figure}[!tp]
    \centering
    \includegraphics[width=0.999\linewidth]{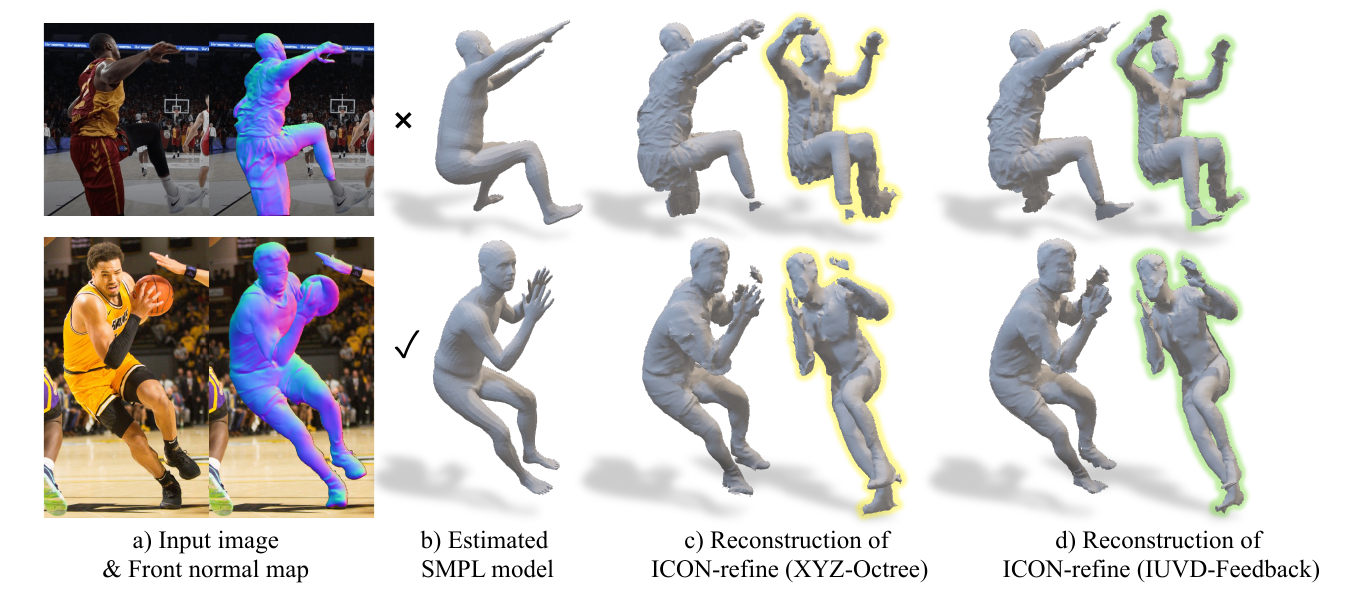}
    \caption{Analysis on the SMPL fitting problem. Given images with severe self-occlusion, the 3D human surface reconstruction results based on inaccurately and accurately estimated SMPL models are shown in the first and second rows, respectively. The highlighted meshes show results from side views.}
    \label{fig:exp_smpl_fit}
\end{figure}

\textit{4) Limitations and future work.} 
The proposed IUVD-Feedback representation has some limitations and requires future work to improve it. Here, we analyze these issues and provide possible research directions for future work.

Firstly, the accuracy of HPS strongly affects our approach, which is a common issue for SMPL-based representations \cite{xiu2022icon,xiu2023econ}. To analyze this issue, we test ICON \cite{xiu2022icon} with the XYZ-Octree and the IUVD-Feedback representations on in-the-wild images with severe self-occlusions. If the estimated SMPL model is not accurate, as shown in the first row of Fig. \ref{fig:exp_smpl_fit}, the final reconstruction will lose accuracy in aspect of human pose but still keep the clothing details in consistent with the estimated normal maps. This is because that the human pose and shape information comes mainly from the SMPL model, but the geometric details come from the normal maps. Thanks to the rapid development of the learning-based human mesh recovery methods \cite{tian2023recovering}, the impact of this issue has been gradually alleviated.

Secondly, upon closer examination of the IUVD-Feedback results in Fig. \ref{fig:exp_qualitative_comparison}, it appears that there is a trade-off between capturing loose clothing details and maintaining a reasonable body shape. This issue arises because the dense correspondence between the IUVD space (\textit{a derivative space of the 2D manifold}) and the XYZ space (\textit{Euclidean space}) is not strictly uniform, particularly when the query points are distant from the body surface. To address this uneven correspondence, the resolution of UVD space could be modified adaptively along the D-axis, i.e. using \textit{dynamic resolution}. By increasing the resolution in regions far from the body space, the details of loose clothing can be better preserved. This suggests a potential improvement for the implicit IUVD representation.

\begin{figure}[!tp]
    \centering
    \includegraphics[width=0.99\linewidth]{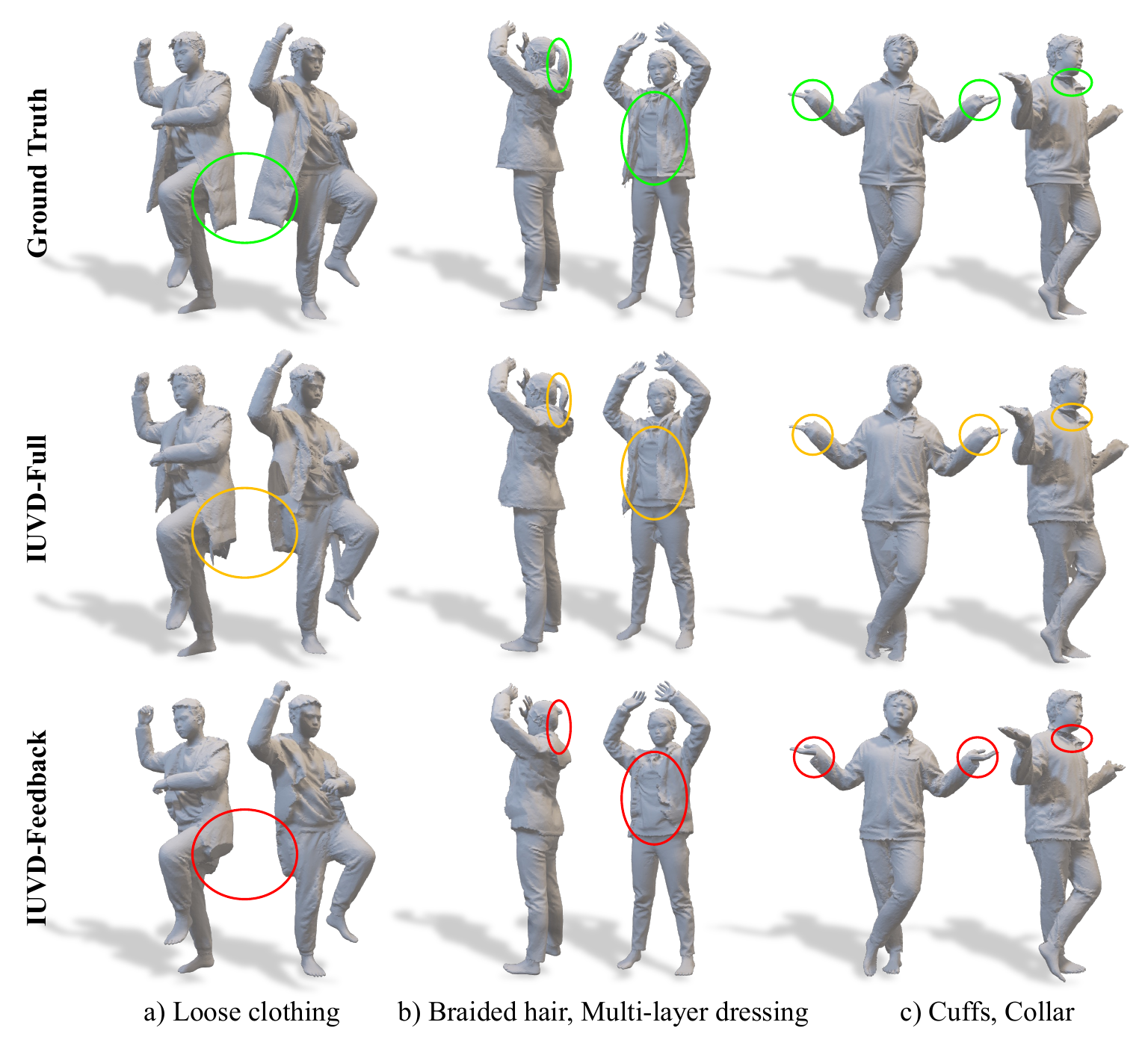}
    \caption{Hard cases for reconstructing the ground-truth scans of THuman2.0 dataset with the proposed IUVD-Full and IUVD-Feedback representations.}
    \label{fig:exp_iuvd_hard_cases}
\end{figure}

Thirdly, there are some hard cases that cannot be fully reconstructed by the IUVD-Feedback representation, including loose clothing, braided hair, cuffs, etc., as shown in the third row of Fig. \ref{fig:exp_iuvd_hard_cases}. Here we provide a possible approach to extend the IUVD-Feedback representation for loose clothing. In the query-and-infer process, we can adaptively extend the range of D-axis or change the query line to a curve, thus removing the convex assumption \ref{assump:2}, which is the main reason for the above problems, and querying more regions. This \textit{adaptive query algorithm} can be guided by clothing semantic segmentation or extrapolated DensePose UV maps to avoid the possible redundancy problem. By retraining the neural networks in IUVD space, the reconstruction accuracy can also be ensured. The expected results are shown in the second row of Fig. \ref{fig:exp_iuvd_hard_cases}, where the resolution of IUVD-Full is set to $24 \times 128 \times 128 \times 41$ for simulating the results of this adaptive query algorithm. It indicates promising research on the implicit IUVD representation in future work.

Fourthly, we hope that the implicit IUVD representation will inspire further research into part-based 3D human surface reconstruction using UV mapping. For example, employing fewer but more meaningful UV segments, such as garment-specific UV maps \cite{jafarian2023normalguided} instead of body part UV maps \cite{guler2018densepose}, could improve the reconstruction continuity and reduce the need for dilation-erosion processing during visualization.

\section{Conclusion} \label{sec:conclusion}
In this paper, we introduced the IUVD-Feedback representation, which comprises a novel implicit function built upon SMPL UV maps and a feedback query algorithm for 3D human surface reconstruction. This representation effectively preserves the pose and shape prior of the SMPL model, and can be flexibly embedded into existing implicit reconstruction pipelines. Based on the designed feature space transformation and the mesh transformation approaches, the implicit function operates within the SMPL-based IUVD space, thereby reducing redundant query points typically encountered in the traditional XYZ space. In IUVD space, the proposed feedback query algorithm further minimizes the redundancy in the implicit query-and-infer process. Experimental results demonstrate that the IUVD-Feedback representation significantly accelerates the query-and-infer and visualization steps of implicit human surface reconstruction while also enhancing the robustness of the reconstructed results. Furthermore, this representation has proven to be applicable to generative tasks by leveraging the semantic information inherent in the parametric body model.


\bibliographystyle{IEEEtran} 
\bibliography{mybib}

\end{document}